\documentclass{sbc2023}%

\usepackage{graphicx}
\usepackage[misc,geometry]{ifsym} 
\usepackage{fontspec}
\usepackage{color}
\usepackage{hyperref} 
\usepackage{aas_macros}
\usepackage[bottom]{footmisc}
\usepackage{pifont}
\usepackage{supertabular}
\usepackage{comment}
\usepackage{afterpage}
\usepackage{float}
\usepackage{url}
\usepackage{booktabs}
\usepackage{pifont}
\usepackage{multicol}
\usepackage{multirow}
\usepackage{array}
\usepackage{ragged2e}
\usepackage{algorithm}
\usepackage{algpseudocode} 
\usepackage{tabularx}
\usepackage{tikz}
\usetikzlibrary{calc, positioning, shapes.geometric} 
\usepackage{amsmath}     
\usepackage{subcaption} 
\usepackage{amssymb}
\usepackage{listings}
\usepackage{xcolor}

\definecolor{codegreen}{rgb}{0,0.6,0}
\definecolor{codegray}{rgb}{0.5,0.5,0.5}
\definecolor{codepurple}{rgb}{0.58,0,0.82}
\definecolor{backcolour}{rgb}{0.95,0.95,0.92}

\lstdefinestyle{mystyle}{
    backgroundcolor=\color{backcolour},   
    commentstyle=\color{codegreen},
    keywordstyle=\color{magenta},
    numberstyle=\tiny\color{codegray},
    stringstyle=\color{codepurple},
    basicstyle=\ttfamily\footnotesize,
    breakatwhitespace=false,         
    breaklines=true,                 
    captionpos=b,                    
    keepspaces=true,                 
    numbers=left,                    
    numbersep=5pt,                  
    showspaces=false,                
    showstringspaces=false,
    showtabs=false,                  
    tabsize=2
}
\setcitestyle{square}

\definecolor{orcidlogo}{rgb}{0.37,0.48,0.13}
\definecolor{unilogo}{rgb}{0.16, 0.26, 0.58}
\definecolor{maillogo}{rgb}{0.58, 0.16, 0.26}
\definecolor{darkblue}{rgb}{0.0,0.0,0.0}
\hypersetup{colorlinks,breaklinks,
            linkcolor=darkblue,urlcolor=darkblue,
            anchorcolor=darkblue,citecolor=darkblue}

\doi{10.5753/jbcs.202X.XXXXXX}

\title[A Comparative Benchmark of Federated Learning Strategies for Mortality Prediction on Heterogeneous and Imbalanced Clinical Data]{
    A Comparative Benchmark of Federated Learning Strategies for Mortality Prediction on Heterogeneous and Imbalanced Clinical Data}

\author[Tertulino 2026]{
\affil{\textbf{Rodrigo Tertulino}~\href{https://orcid.org/0000-0002-7594-9312}{\textcolor{orcidlogo}{\textbf{\small ORCID}}}~~[~\textbf{IFRN}~|\href{mailto:rodrigo.tertulino@ifrn.edu.br}{~\textbf{\textit{rodrigo.tertulino@ifrn.edu.br}}}~]}

}

\usepackage{placeins}
\begin{document}

\begin{frontmatter}
\maketitle

\begin{mail}
Federal Institute of Education, Science, and Technology of Rio Grande do Norte (IFRN), Software Engineering and Automation Research Laboratory - LaPEA, 59628-330, Mossoró-RN, Brazil.
\end{mail}



\begin{abstract}
\textbf{Abstract.~}
\noindent Machine learning models hold significant potential for predicting in-hospital mortality, yet data privacy constraints and the statistical heterogeneity of real-world clinical data often hamper their development. Federated Learning (FL) offers a privacy-preserving solution; however, its performance under non-independent and Identically Distributed (non-IID) and imbalanced conditions requires rigorous investigation. The study presents a comparative benchmark of five federated learning strategies: FedAvg, FedProx, FedAdagrad, FedAdam, and FedCluster for mortality prediction. Using the large-scale MIMIC-IV dataset, we simulate a realistic non-IID environment by partitioning 466,351 admissions across five clinical care units, one of which (Labor \& Delivery) contains a single mortality case in over 20,000 admissions, and enrich the feature set with an 11-item first-24-hour laboratory panel alongside demographics and utilization counts. Following the precedent of Harutyunyan et al.'s MIMIC-III clinical benchmark suite, whose closest-matched task has an almost identical 2.06\% positive rate to our own 1.98\%, we report AUC-ROC and AUC-PR as primary, threshold-independent metrics rather than F1, which is unreliable at this level of class imbalance. Our experiments, conducted over 50 communication rounds and five random seeds, show that the regularization-based strategy, FedProx, achieved the best AUC-ROC (0.897, mean over seeds) and the best mean AUC-PR (0.230) among the five FL strategies -- ahead of FedAdagrad (0.890/0.202/0.232), FedAdam (0.885/0.192/0.270), FedCluster (0.873/0.221/0.280), and FedAvg (0.857/0.177/0.255) -- with paired $t$-tests confirming that its AUC-ROC advantage is statistically significant against every other strategy; on the fixed-threshold F1-score, however, FedProx (0.273) is narrowly surpassed by FedCluster (0.280), so no single strategy dominates every metric. The best centralized baseline (naive, no explicit imbalance handling) achieves AUC-ROC of 0.929 and AUC-PR of 0.312, significantly outperforming FedProx on both. Breaking performance down by client, following Harutyunyan et al.'s per-phenotype table, reveals that the federated global model does not serve all care units equally: AUC-ROC ranges from 0.809 to 0.902 across the evaluable clients, with the smallest, most clinically distinct clients faring worst under aggregation dominated by the largest client's training volume. Our findings indicate that regularization-based FL algorithms like FedProx remain the more robust choice among federated aggregation strategies for heterogeneous and imbalanced clinical prediction tasks, while centralization, where permissible, still holds a slight predictive edge, and per-client heterogeneity in outcomes deserves attention independent of the aggregate numbers. The work provides an empirical benchmark and a methodological cautionary note on evaluation-set construction and metric choice for selecting appropriate FL strategies for real-world healthcare applications.
\end{abstract}



\end{frontmatter}

\section{Introduction}
\label{sec:intro}

Predicting patient outcomes in Intensive Care Units (ICUs) represents a critical area of clinical research, where timely and accurate forecasts can significantly influence therapeutic decisions and resource allocation \cite{10.1093/ehjacc/zuae037}. Machine learning (ML) and deep learning (DL) models have demonstrated considerable promise in the domain, leveraging the vast amount of data available in Electronic Health Records (EHRs) to identify complex patterns indicative of patient deterioration and mortality risk \cite{10.3389/fmed.2019.00036}. However, the full potential of these models is often constrained by a fundamental challenge: the sensitive nature of medical data.

Training high-performance ML models typically requires large, diverse datasets. However, clinical data are subject to stringent privacy regulations, such as the General Data Protection Regulation (GDPR)~\cite{10.1145/3675888.3676142} and the Health Insurance Portability and Accountability Act (HIPAA)~\cite{10.1145/1273353.1273354}. Consequently, patient information remains siloed within individual healthcare institutions, creating significant barriers to collaborative research and model development. Centralizing multi-institutional data for training is logistically complex and poses substantial privacy and security risks \cite{Mothukuri2021}.

To overcome these barriers, Federated Learning (FL) has emerged as a transformative paradigm for privacy-preserving collaborative machine learning \cite{DBLP:journals/corr/McMahanMRA16}. The FL framework enables training a global model across multiple decentralized clients, such as hospitals, without requiring the sharing of raw patient data. Instead, each client trains a local model on its data, and only the resulting model updates (e.g., weights or gradients) are sent to a central server for aggregation~\cite{10.1145/3594300.3594312}. Through an iterative process, a robust global model is collaboratively built, with sensitive information never leaving the institutional firewalls. The utility of FL has already been demonstrated across various medical applications, including sepsis prediction \cite{Ding2025-ci}, cardiovascular event forecasting \cite{ZHANG2021103009}, and medical imaging analysis \cite{Kaissis2020, Starke2025}.

Despite its advantages, the practical implementation of FL in healthcare is confronted by statistical heterogeneity. Data distributions among different hospitals are frequently not independent and identically distributed (non-IID), owing to variations in patient demographics, clinical practices, and medical equipment \cite{YURDEM2024e38137, electronics14091750}. Standard FL algorithms, most notably Federated Averaging (FedAvg), can experience significant performance degradation, slower convergence, or even divergence when faced with non-IID data \cite{Khowaja2023}. Moreover, clinical prediction tasks, such as mortality prediction, are often characterized by severe class imbalance, a problem that is compounded by the Non-IID nature of federated datasets.

Various advanced FL strategies have been developed to enhance model performance and stability in heterogeneous environments in response to these issues. While numerous studies have applied FL to clinical prediction tasks \cite{Dayan2025, Chen2025ICUMortality, Zhang2025}, a comprehensive, empirical comparison of modern aggregation strategies remains an open area of investigation. Specifically, there is a need to evaluate how algorithms employing adaptive server-side optimizers (e.g., FedAdam, FedAdagrad) or client-side regularization (e.g., FedProx) perform in comparison to clustering-based approaches on a complex, real-world clinical dataset for mortality prediction.

This work aims to address the gap by conducting a rigorous comparative analysis of five distinct federated learning strategies for predicting in-hospital mortality. Leveraging the large-scale MIMIC-IV critical care dataset, we simulate a realistic non-IID federated environment by partitioning data based on hospital care units~\cite{Johnson2024-on}. The \texttt{Synthetic Minority Over-sampling Technique with Tomek Links (SMOTETomek)} resampling technique is applied at the client level to tackle the issue of class imbalance inherent in mortality prediction~\cite{Khleel2023-ef}. The core contributions of our study are: 

\begin{itemize}

\item A comprehensive performance evaluation of FedAvg, FedProx, FedAdagrad, FedAdam, and FedCluster in a realistic clinical setting.

\item An analysis of each strategy's convergence speed, stability, and computational efficiency. 

\item Actionable insights into the most suitable FL algorithms for privacy-preserving mortality prediction tasks on heterogeneous and imbalanced data.

\end{itemize}

The remainder of the paper is organized as follows. Section 2 reviews related work in federated learning for healthcare. Section 3 details the dataset, data preprocessing steps, and the specifics of the five federated learning strategies evaluated. Section 4 presents the experimental setup and the results of our comparative analysis. Section 5 provides a discussion of the findings, and finally, Section 6 concludes the paper with a summary of our contributions and directions for future work.

\section{Related Work}
\label{sec:related_work}

Federated Learning (FL) is being increasingly adopted as a fundamental technology for predictive modeling in healthcare, enabling multi-institutional collaboration while preserving patient privacy \cite{Mothukuri2021}. A growing body of literature demonstrates its successful application across a range of clinical tasks, from heart disease prediction using benchmark datasets \cite{S2022} to more complex applications, such as heart sound classification from audio signals \cite{Qiu2022} and medical imaging analysis \cite{Starke2025, Speth2025}. Within the critical care domain, several studies have specifically leveraged the MIMIC family of datasets to develop and validate FL frameworks for mortality prediction, underscoring the relevance of the research area.

For instance, \citet{Chen2025ICUMortality} introduced FedHealth-Net, a federated transfer learning framework for ICU mortality prediction using the MIMIC-III and eICU datasets. Their approach focused on knowledge transfer to improve model performance on clients with smaller datasets. Similarly, \citet{AlTaei2025} employed a standard FedAvg approach on the MIMIC-IV dataset to predict major adverse cardiovascular events, successfully demonstrating the feasibility of FL for complex predictive tasks. Expanding on these applications, \citet{Zhang2025} demonstrated the utility of FL in predicting postoperative delirium across a multi-center EHR dataset, further confirming the viability of the baseline FedAvg algorithm for clinical predictive tasks. Moreover, \citet{Dayan2025} conducted a large-scale study using real-world EHR data from 176 hospitals to predict clinical outcomes, providing strong evidence of FL's applicability in highly heterogeneous environments.

Despite these significant contributions, several gaps persist in the existing literature. A primary weakness is that many studies focus on demonstrating the viability of FL using a single baseline algorithm, typically FedAvg \cite{AlTaei2025, Lee2025, S2022}. While some benchmark studies have compared foundational algorithms like FedAvg and FedProx \cite{10.1145/3286490.3286559}, these evaluations are often conducted on non-clinical benchmark datasets and do not address the combined challenges of clinical data heterogeneity and severe class imbalance. Consequently, there is limited understanding of how more advanced algorithms, such as server-side adaptive optimizers or clustering-based approaches, perform relative to one another. In a similar critical care context, \citet{10.1145/3514500} presented an important benchmark for mortality prediction on the eICU dataset, evaluating and comparing FedAvg and FedProx and concluding that FedProx offered performance benefits in a heterogeneous setting.

Another significant limitation is the approach to data-related challenges. While the non-IID nature of medical data is widely acknowledged \cite{Dayan2025, Starke2025}, the combined problem of severe class imbalance alongside the heterogeneity is often not explicitly addressed. Many studies rely on implicit aggregation to address these issues rather than incorporating dedicated client-level balancing techniques, such as the SMOTE-Tomek method used in our work. As summarized in Table~\ref{tab:related_work_summary}, existing research has laid a crucial foundation, yet a comprehensive benchmark evaluating a diverse set of modern FL algorithms on a large-scale critical care dataset for mortality prediction remains absent.

Therefore, our study aims to bridge these identified gaps. We provide a rigorous, head-to-head comparison of five distinct FL strategies spanning baseline, regularized, adaptive, and clustering-based approaches on the MIMIC-IV dataset. By explicitly addressing class imbalance at the client level, we evaluate these algorithms under conditions that more closely mirror real-world clinical data challenges, offering a clearer perspective on their relative strengths and weaknesses for high-stakes predictive tasks.

\begin{table*}[t] 
\centering
\caption{Summary and comparison of relevant studies in Federated Learning for healthcare.}
\label{tab:related_work_summary}
\renewcommand{\arraystretch}{1.3} 
\begin{tabularx}{\textwidth}{@{} l >{\raggedright\arraybackslash}X >{\raggedright\arraybackslash}p{2.5cm} >{\raggedright\arraybackslash}p{3cm} >{\justifying\arraybackslash}X @{}}
\toprule
\textbf{Reference} & \textbf{Objective} & \textbf{Dataset(s)} & \textbf{FL Approach} & \textbf{Identified Gap / Weakness} \\ \midrule

\cite{AlTaei2025} & Predict major adverse cardiovascular events & MIMIC-IV & FedAvg & Limited to a single baseline FL algorithm (FedAvg). Does not explicitly address class imbalance. \\ 

\cite{Chen2025ICUMortality} & Predict ICU mortality with privacy & MIMIC-III, eICU & FedHealth-Net (Transfer Learning) & Focuses on transfer learning rather than a broad comparison of aggregation optimizers. \\ 

\cite{Dayan2025} & Predict clinical outcomes in a real-world ICU setting & HCA Healthcare EHR & FedAvg, FedProx & Comparison limited to FedAvg and FedProx. Does not investigate adaptive optimizers. \\ 

\cite{Lee2025} & Predict sepsis using data from two institutions & Private EHR & FedAvg & Small number of clients (2). Limited to FedAvg; no discussion of non-IID issues. \\ 

\cite{S2022} & Predict heart disease & UCI Heart Disease & FedAvg & Uses a non-clinical benchmark dataset. Comparison limited to different models under FedAvg. \\ 

\cite{Qiu2022} & Classify heart sounds & PhysioNet/CinC Challenge 2016 & Federated Transfer-Meta Learning & Focuses on a different data modality (audio signals). Not directly comparable to EHR-based prediction. \\ 

\cite{Zhang2025} & Predict postoperative delirium in older adults & Multi-center EHR & FedAvg & Application-specific; comparison limited to FedAvg and a centralized model. \\ 

\cite{Starke2025} & Cardiac MRI segmentation & Multi-center Cardiac MRI & FedAvg, FedProx, FedCurv & Focuses on medical imaging (segmentation task). Different data modality and objective. \\ 

\cite{10.1145/3514500} & Benchmark FL algorithms for mortality prediction  & eICU & FedAvg, FedProx, SCAFFOLD & Does not evaluate server-side adaptive optimizers. Does not explicitly address class imbalance. \\ 

\cite{10.1145/3286490.3286559} & Benchmark FL algorithms against a centralized approach & MNIST & FedAvg, FedSGD, FedProx & Uses a non-clinical, balanced image dataset. Lacks comparison with adaptive optimizers. \\ \bottomrule
\end{tabularx}
\end{table*}

\section{Methodology}
\label{sec:methodology}

The section details the comprehensive methodology employed in our study, encompassing the dataset and cohort selection, data preprocessing techniques, the federated learning simulation environment, the evaluated models and aggregation strategies, and the metrics used for performance evaluation.

\subsection{Dataset and Cohort Selection}
The study utilizes the \texttt{hosp} module of the Medical Information Mart for Intensive Care (MIMIC-IV) database (version 3.1). The data was installed and managed in a PostgreSQL database for efficient querying and processing. For our analysis, a specific cohort was defined and accessed via a materialized view, \texttt{mimiciv\_hosp.mortalidade\_features}, which contains pre-joined and filtered data for the binary classification task of predicting in-hospital mortality.

The study population was defined by specific inclusion and exclusion criteria to ensure a relevant and consistent cohort. The inclusion criteria for our study were as follows: adult patients \texttt{(age $\geq$ 18 years)} admitted to the hospital. The analysis retained all qualifying hospital admissions rather than only a patient's first admission; because the cohort contains multiple admissions for some patients (546,028 admissions from approximately 223,000 distinct patients), the risk of leakage and correlation from repeated encounters is instead controlled at the split stage, by partitioning the data at the patient level so that all admissions belonging to the same patient are kept together in a single train, validation, or test partition (Section~\ref{sec:non_iid_partitioning}). Furthermore, admissions with missing data for the primary outcome variable (\texttt{hospital\_expire\_flag}) were excluded.

The selection process resulted in a final cohort of 546,028 unique hospital admissions. The features selected for the model include patient demographics, insurance type, admission-level diagnosis/procedure counts, and a first-24-hour laboratory panel (Section~\ref{sec:preprocessing} details all eleven measurements and how missingness is handled). To apply for access to the database, the author (R. Tertulino) passed the examination on the protection of human research participants and obtained the certificate (No. 71352117).

The baseline demographic and clinical characteristics of the final study cohort are presented in Table~\ref{tab:demographics}. The cohort consists predominantly of adult patients, with a mean age and gender and race distributions consistent with those of a large tertiary care center in the United States. The median length of hospital stay was 2.8 days (Interquartile Range: 1.1 to 5.6 days). The overall in-hospital mortality rate was 2.16\% across this full cohort; within the five-care-unit subset used for all federated and centralized experiments (Section~\ref{sec:non_iid_partitioning}), it is 1.98\%.

\begin{table}[h!]
\centering
\caption{Baseline demographic and clinical characteristics of the study cohort.}
\label{tab:demographics}
\begin{tabular}{@{}ll@{}}
\toprule
\textbf{Characteristic} & \textbf{Value (N = 546,028)} \\ \midrule
\textbf{Age (years)} & \\
\multicolumn{1}{l}{\quad Mean (SD)} & 56.9 $\pm$ 19.0 \\
\multicolumn{1}{l}{\quad Median (IQR)} & 58.0 (43.0 - 71.0) \\
 & \\
\textbf{Gender, n (\%)} & \\
\multicolumn{1}{l}{\quad Female} & 284,097 (52.0\%) \\
\multicolumn{1}{l}{\quad Male} & 261,931 (48.0\%) \\
 & \\
\textbf{Race, n (\%)} & \\
\multicolumn{1}{l}{\quad White} & 336,538 (61.6\%) \\
\multicolumn{1}{l}{\quad Other} & 126,199 (23.1\%) \\
\multicolumn{1}{l}{\quad Black/African American} & 75,482 (13.8\%) \\
\multicolumn{1}{l}{\quad Asian} & 7,809 (1.4\%) \\
 & \\
\textbf{Length of Stay (days)} & \\
\multicolumn{1}{l}{\quad Mean (SD)} & 4.8 $\pm$ 7.2 \\
\multicolumn{1}{l}{\quad Median (IQR)} & 2.8 (1.1 - 5.6) \\
 & \\
\textbf{In-Hospital Mortality, n (\%)} & \\
\multicolumn{1}{l}{\quad Survived (0)} & 534,227 (97.8\%) \\
\multicolumn{1}{l}{\quad Died (1)} & 11,801 (2.2\%) \\ \bottomrule
\end{tabular}
\end{table}

\subsection{Data Preprocessing and Feature Engineering}
\label{sec:preprocessing}
The feature set combines three groups of variables: (1) demographics and administrative fields -- age, gender, race, and insurance type; (2) two coarse utilization proxies -- the total number of ICD diagnosis and procedure codes assigned during the admission; and (3) an 11-item laboratory panel drawn from \texttt{mimiciv\_hosp.labevents} -- creatinine, blood urea nitrogen, sodium, potassium, glucose, white blood cell count, hemoglobin, hematocrit, platelet count, bicarbonate, and lactate -- each averaged over the first 24 hours following admission (\texttt{admittime}, from \texttt{mimiciv\_hosp.admissions}), which is the only place in the pipeline where a value's timestamp is used to enforce an admission-time-only feature window and avoid look-ahead leakage from labs drawn later in a stay, closer to the outcome. \texttt{mimiciv\_icu.chartevents} vital signs (heart rate, blood pressure, etc.) were considered but excluded: they require an ICU stay, and only 9-21\% of admissions in the five selected care units have one (Section~\ref{sec:non_iid_partitioning}), since none of the five largest care units by volume are themselves ICUs; \texttt{labevents}, ordered for any hospitalized patient regardless of unit, has 63-68\% coverage for ten of the eleven panel items (lactate is the exception, at 12\%, reflecting that it is selectively ordered for more acutely ill patients rather than missing at random).

\begin{figure}[ht]
    \centering
    \includegraphics[width=\columnwidth]{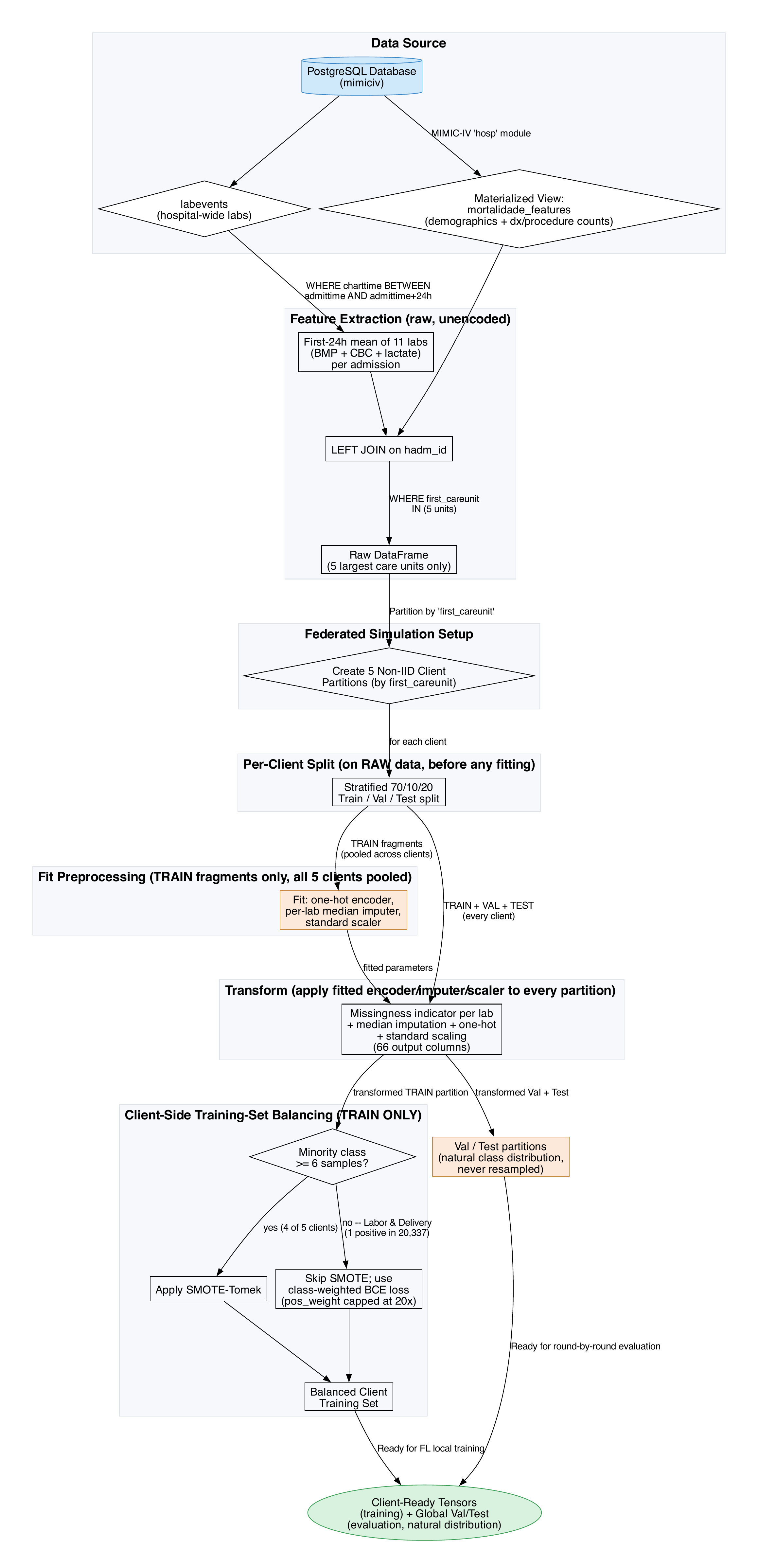}
    \caption{The end-to-end data preprocessing pipeline, from data extraction and laboratory-panel enrichment in the PostgreSQL database, through the per-client train/validation/test split, to the creation of balanced, non-IID training sets for each client in the federated network. The encoder, per-lab median imputer, and feature scaler are fit only on the pooled training fragments across all five clients (never on validation/test rows), and SMOTE-Tomek is likewise applied to each client's training partition only -- validation and test data pass through the fitted transform unresampled, preserving each client's natural class distribution end to end.}
    \label{fig:preprocessing_pipeline}
\end{figure}

Categorical features were converted to numerical values using one-hot encoding. Because roughly a third of admissions lack a value for any given lab within the first 24 hours, each lab feature is paired with a binary missingness indicator (fit and applied identically to the SMOTE-Tomek pipeline described in Section~\ref{sec:non_iid_partitioning}: computed on pooled training data only, never leaking validation/test information), and missing values are imputed with the training-set median before the numeric features are standardized. All features were subsequently cast to 32-bit floating-point to ensure compatibility with deep learning frameworks. The final feature matrix (66 columns after one-hot encoding and the addition of missingness indicators) was used to predict the binary outcome, \texttt{hospital\_expire\_flag}. A comprehensive overview of the entire data preparation pipeline is illustrated in Figure~\ref{fig:preprocessing_pipeline}.

\subsection{Federated Learning Simulation Setup}
To evaluate the performance of FL strategies under realistic conditions, we simulated a cross-silo federated environment characterized by statistical heterogeneity.

\subsubsection{Non-IID Data Partitioning}
\label{sec:non_iid_partitioning}
Real-world medical data is inherently non-IID because patient populations and clinical practices vary across institutions. The entire dataset was partitioned into distinct client datasets to simulate the heterogeneity based on the \texttt{first\_careunit} feature. Each unique care unit represents a client in the federation. For our experiments, we selected the five largest care units to act as the five clients in our federated network: \texttt{Emergency Department, Discharge Lounge, Labor \& Delivery, Medicine/Cardiology, and Medicine}. 

Together, the five selected care units account for 466,351 of the cohort's 546,028 admissions (85.4\%); the remaining admissions, spread across smaller care units, are not used in either the federated or the centralized experiments, so that both paradigms are compared on the identical underlying population. The resulting data distribution across these five clients is detailed in Table~\ref{tab:data_partitioning}. The table reveals a significant skew in the number of samples per client and an extreme variance in the label distribution. The mortality rate ranges from nearly zero in the 'Labor \& Delivery' unit to almost 3\% in the 'Medicine' unit. The class imbalance at the client level thus presents a considerable challenge for standard aggregation algorithms, such as FedAvg: in such a scenario, the updates from clients with large, heavily imbalanced datasets can overwhelm the contributions from smaller clients that may hold more informative patterns regarding the minority class. This confirms the strong non-IID nature of our experimental setup.

\begin{table*}[h!]
\centering
\caption{Data distribution and characteristics across the five clients after non-IID partitioning. The table displays the total number of samples (hospital admissions) and the class distribution for the mortality prediction task.}
\label{tab:data_partitioning}
\begin{tabular}{@{}lcccc@{}}
\toprule
\textbf{Client (Care Unit)} & \textbf{Total Samples} & \textbf{Survivors (Class 0)} & \textbf{Mortality (Class 1)} & \textbf{Mortality Rate (\%)} \\ \midrule
Emergency Department & 374,252 & 365,678 & 8,574 & 2.29 \\
Discharge Lounge     & 49,386  & 49,218  & 168   & 0.34 \\
Labor \& Delivery    & 20,337  & 20,336  & 1     & 0.00 \\
Medicine/Cardiology  & 11,209  & 11,060  & 149   & 1.33 \\
Medicine             & 11,167  & 10,833  & 334   & 2.99 \\ \bottomrule
\end{tabular}
\end{table*}

\subsubsection{Train/Validation/Test Split}
The split is performed at the \emph{patient} level to prevent leakage from patients with more than one admission: every patient (\texttt{subject\_id}) is assigned in its entirety to one of a 70\% training, 10\% validation, or 20\% test partition -- stratified by whether the patient died in any of their admissions, so the overall mortality prevalence is preserved across the three partitions -- so that no patient's records ever appear in more than one partition. Each client's (care unit's) local training, validation, and test partitions are then the subsets of that unit's admissions whose patients fall in the corresponding global partition, before any class-imbalance handling is applied. The five clients' test subsets are concatenated into a single global test set, and their validation subsets into a single global validation set, both used to evaluate and calibrate every strategy -- federated and centralized alike -- so that all reported comparisons share exactly the same held-out data. Both the global validation and global test sets retain each client's natural mortality rate; only training partitions are ever resampled.

\subsubsection{Client-Side Class Imbalance Handling}
In-hospital mortality prediction is a task characterized by severe class imbalance, as the number of surviving patients far exceeds the number of non-survivors. To address the issue at the client level without centralized data access, we applied the Synthetic Minority Over-sampling Technique combined with Tomek Links (SMOTE-Tomek)~\cite{Khleel2023-ef}. The hybrid method was applied locally to each client's training data partition before the start of local training epochs, and only to the training partition: each client's validation and test partitions retain their natural, unresampled class distributions, so evaluation always reflects the true mortality rate. The technique works by oversampling the minority class (mortality) through synthetic data generation and cleaning the feature space by removing Tomek links, resulting in a more balanced and well-defined training set for each client.

One client, Labor \& Delivery, is a degenerate case: it has only a single mortality event across all 20,337 admissions (Table~\ref{tab:data_partitioning}). Under the patient-level split, that one positive patient falls entirely into a single partition, so the client's training partition contains at most one positive example -- below the minimum number of minority samples SMOTE's nearest-neighbor search requires (at least $k_{\text{neighbors}}+1$, conventionally 6). SMOTE-Tomek is therefore skipped for this client, which instead relies solely on class-weighted binary cross-entropy loss. The positive-class weight is capped at 20$\times$ (applied identically, on top of SMOTE-Tomek, for every client) to prevent an extreme per-client imbalance ratio from letting one client's gradient dominate and destabilize the federated aggregation step.

\subsection{Federated Learning Model}
Every FL strategy and centralized baseline in this study shares a single model architecture, a Deep Neural Network (DNN), so that the comparison across aggregation strategies is not confounded by also varying the underlying model. The DNN is a multi-layer perceptron designed to capture non-linear relationships in the data: an input layer, two hidden layers with 128 and 64 neurons respectively, each followed by the Rectified Linear Unit (ReLU) activation function, and a final output layer with a single neuron and a sigmoid activation function for binary classification.

\subsection{Evaluated Federated Learning Strategies}
\label{sec:fl_strategies}
A core objective of the work is to compare the performance of different FL aggregation algorithms. We evaluated five distinct strategies:

\begin{itemize}
    \item \textbf{Federated Averaging (FedAvg):} The foundational FL algorithm that aggregates client models by computing a weighted average of their parameters~\cite{mcmahan2017communication}. The global model $w^{t+1}$ at round $t+1$ is updated as follows:
    $$ w^{t+1} = \sum_{k=1}^{K} \frac{n_k}{N} w_k^{t+1} $$
    where $K$ is the number of clients, $w_k^{t+1}$ represents the model parameters from client $k$ after local training, $n_k$ is the size of the dataset on client $k$, and $N$ is the total number of data samples across all clients.
    
    \item \textbf{FedProx:} An extension of FedAvg designed to improve stability in non-IID settings~\cite{electronics12204364}. It introduces a proximal term into the local client's objective function, which prevents local updates from diverging too far from the global model \cite{Dayan2025}. The local objective function for a client $k$ is modified as:
    $$ F_k(w) = L_k(w) + \frac{\mu}{2} \|w - w^t\|^2 $$
    where $L_k(w)$ is the local loss, $w^t$ is the global model from round $t$, and $\mu$ is a hyperparameter controlling the regularization strength. The aggregation step remains the same as in FedAvg.
    
    \item \textbf{FedAdagrad \& FedAdam:} These strategies adapt the principles of the Adagrad and Adam optimizers to the server-side aggregation process, maintaining adaptive learning rates for the global model parameters~\cite{reddi2021adaptivefederatedoptimization}. Let $\bar{w}_t = \sum_{k=1}^{K} \frac{n_k}{N} w_k^{t+1}$ be the FedAvg-style client average at round $t$, and $\Delta_t = \bar{w}_t - w^t$ the resulting pseudo-gradient (the direction, and magnitude, of the average client update). The global model $w^t$ is updated according to the following rules:
    \begin{itemize}
        \item For \textbf{FedAdagrad}, the update rule is:
        $$ w^{t+1} = w^t + \eta \frac{\Delta_t}{\sqrt{v_t} + \tau} \quad \text{where} \quad v_t = v_{t-1} + \Delta_t^2 $$
        \item For \textbf{FedAdam}, the update rule incorporates momentum ($m_t$) and velocity ($v_t$) terms. The global model is updated as follows:
    \begin{align*}
        w^{t+1} &= w^t + \eta \frac{m_t}{\sqrt{v_t} + \tau} \\
        \text{where} \quad m_t &= \beta_1 m_{t-1} + (1 - \beta_1) \Delta_t \\
        v_t &= \beta_2 v_{t-1} + (1 - \beta_2) \Delta_t^2
    \end{align*}
    \end{itemize}
    Here, $\eta$ is the server learning rate, $\tau$ is a small constant for numerical stability, and $\beta_1$ and $\beta_2$ are the momentum decay factors ($\beta_1=0.9$, $\beta_2=0.99$ in our experiments).

    \item \textbf{FedCluster:} An approach that aims to handle highly divergent client updates~\cite{10.1145/3579654.3579732}. The aggregation is a multi-step process. First, the model updates from all clients, $\delta_k^t = w_k^{t+1} - w^t$, are calculated and L2-normalized to unit length, so that DBSCAN clusters clients by the \emph{direction} of client drift rather than by update magnitude (which otherwise dominates Euclidean distance in a high-dimensional parameter space and would defeat the clustering). These normalized updates are then clustered using the DBSCAN algorithm to identify a consensus group. The final global model is updated by averaging only the (un-normalized) updates from the largest identified cluster, $C^*_t$, effectively filtering out outliers:
    $$ w^{t+1} = w^t + \frac{1}{|C^*_t|} \sum_{\delta_k^t \in C^*_t} \delta_k^t $$
    With only 5 clients per round, DBSCAN's neighborhood radius $\varepsilon$ is set per round as the median pairwise distance among the normalized updates, rather than a single fixed value, so the notion of "consensus" adapts to how spread out that round's updates are.
\end{itemize}

\subsection{Experimental Parameters and Evaluation Metrics}
All experiments used a consistent set of hyperparameters to ensure a fair comparison, as detailed in Table~\ref{tab:hyperparameters}. The local client models were trained using the Adam optimizer. The client batch size (512) was chosen for computational tractability: at a smaller batch size, one communication round of local training on the Emergency Department client alone (approximately 510,000 post-SMOTE-Tomek rows $\times$ 3 local epochs) would require tens of thousands of gradient steps, making the full 50-round $\times$ 5-strategy sweep computationally impractical on commodity hardware; no other hyperparameter was altered.

\begin{table}[h]
\centering
\caption{Hyperparameters used for all federated learning experiments.}
\label{tab:hyperparameters}
\begin{tabular}{@{}ll@{}}
\toprule
\textbf{Parameter} & \textbf{Value} \\ \midrule
Number of Clients & 5 \\
Total Communication Rounds & 50 \\
Local Training Epochs & 3 \\
Client Batch Size & 512 \\
Local Optimizer & Adam \\
Local Learning Rate & 0.001 \\ \bottomrule
\end{tabular}
\end{table}

For FedAdagrad and FedAdam, the server-side learning rate was $\eta=0.01$ and the numerical-stability constant $\tau=10^{-3}$; for FedProx, the proximal term coefficient was $\mu=0.1$.

\subsubsection{Implementation}
The experimental framework is built on PyTorch for neural network modeling, pandas and scikit-learn for data manipulation, imbalanced-learn for SMOTE-Tomek, and Opacus for per-example DP-SGD (Section~\ref{sec:dp_cost}). Federated orchestration is a plain synchronous Python loop rather than a framework such as Flower: for 5 simulated clients, a full actor-based simulation framework adds process-management overhead without a corresponding benefit, and it makes the custom FedCluster aggregation rule (not available in standard FL frameworks) straightforward to implement and unit-test in isolation. The complete implementation -- data extraction and non-IID partitioning, all five aggregation strategies, the centralized baselines, and the DP-SGD experiment -- is publicly available at \url{https://github.com/rodrigoronner/FedMIMIC-Mortality-Benchmark} to allow independent reproduction of every result reported in this paper.

\subsubsection{Evaluation Metrics}
\label{sec:eval_metrics}
Our primary evaluation metrics are the area under the receiver operating characteristic curve (AUC-ROC) and the area under the precision-recall curve (AUC-PR, equivalently, average precision). Both are threshold-independent: they summarize a classifier's ranking quality across all possible decision thresholds rather than committing to one, which matters here because, at the roughly 2\% mortality prevalence in our experimental cohort (1.98\%), no single fixed threshold is representative of a model's discriminative ability. This choice follows the closest methodological precedent for this exact problem: the widely-used MIMIC-III clinical benchmark suite of \citet{harutyunyan2019multitask} reports AUC-ROC as its primary metric and AUC-PR as a secondary metric for its in-hospital-mortality and physiologic-decompensation tasks specifically \emph{because} F1-style fixed-threshold metrics are unreliable under skew (their decompensation task has a 2.06\% positive rate -- within 0.1 percentage points of ours); the same trade-off is formalized by \citet{davis2006relationship}, who show that AUC-PR is more informative than AUC-ROC precisely when the positive class is rare, since AUC-ROC can look deceptively good even for classifiers with poor precision at realistic operating points. We report AUC-ROC/AUC-PR at every communication round (computed on the full probability output, not a thresholded prediction) to track convergence, and as the headline final numbers in Table~\ref{tab:final_results}.

We additionally report Accuracy, Precision, Recall, and F1-Score, together with the test-set binary cross-entropy loss, as secondary, threshold-dependent metrics -- both at a fixed 0.5 threshold for the round-by-round trajectories in Figures~\ref{fig:performance_evolution} and~\ref{fig:secondary_evolution}, and, after the final round, at a validation-calibrated threshold (a grid search over 99 candidate thresholds, 0.01 to 0.99 in steps of 0.01, on the global validation partition, selecting the threshold that maximizes F1) for the final comparison in Table~\ref{tab:final_results}. These are included because they remain widely reported in the applied clinical FL literature and are more directly interpretable by a non-specialist audience, but given the prevalence of this task, they should be read as secondary to AUC-ROC/AUC-PR, not as the primary basis for comparing strategies.

To quantify run-to-run variability rather than report a single point estimate, every strategy -- federated and centralized -- was trained and evaluated across five independent random seeds, each of which redraws the patient-level split and reinitializes the model. Final metrics are reported as mean $\pm$ standard deviation across these five seeds (Tables~\ref{tab:final_results} and~\ref{tab:final_results_centralized}). Because all strategies within a given seed are evaluated on that seed's identical held-out test partition, differences between strategies are assessed with two-sided paired $t$-tests (paired by seed); we report these for the best federated strategy against every other strategy on the two primary metrics (Table~\ref{tab:significance}). The round-by-round evolution figures and the per-client and differential-privacy analyses use a single representative seed, since their purpose is to illustrate learning dynamics and heterogeneity rather than to rank strategies.

\section{Results}
\label{sec:results}

The section presents the empirical results of our comparative analysis of five federated learning strategies. We evaluate the strategies based on their performance evolution across communication rounds, their final predictive performance, and their computational efficiency.

\subsection{Convergence and Performance Evolution}
\begin{figure*}[ht]
    \centering
    \begin{subfigure}[b]{0.48\textwidth}
        \centering
        \includegraphics[width=\textwidth]{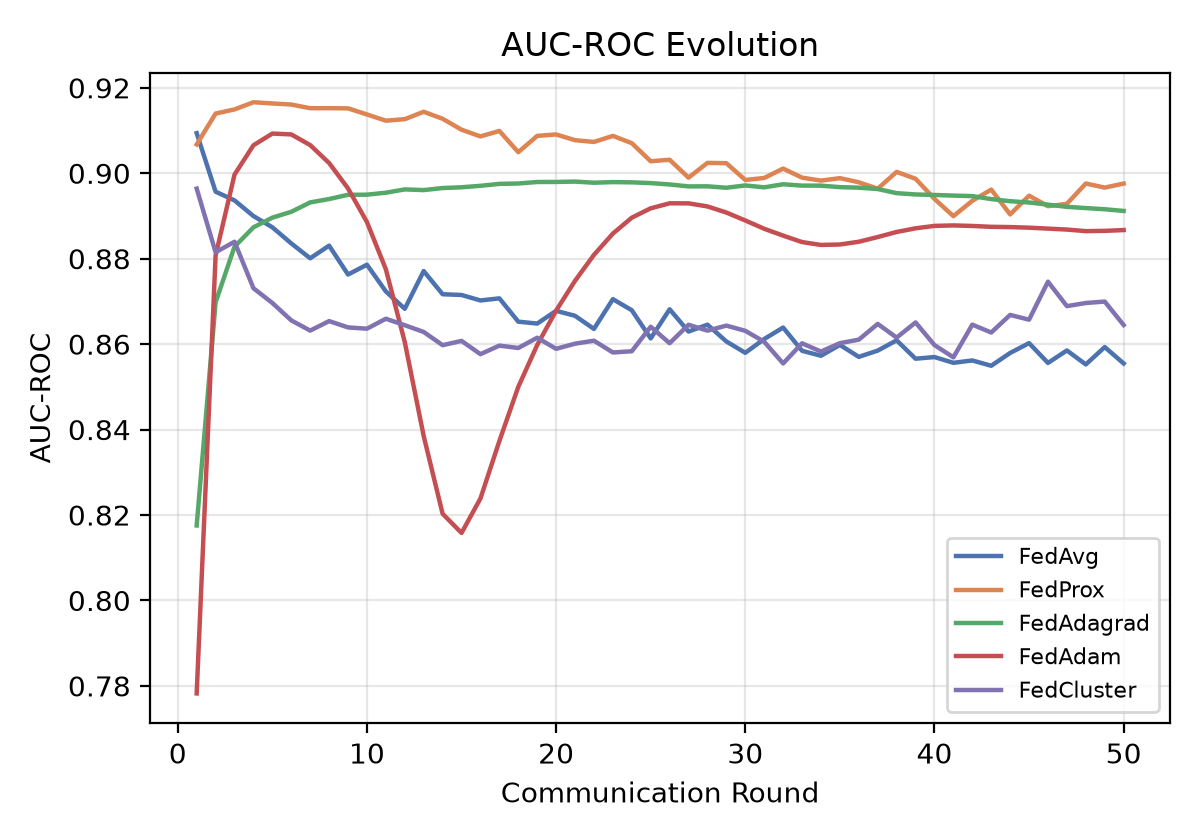}
        \caption{AUC-ROC Evolution}
        \label{fig:evo_auc_roc}
    \end{subfigure}
    \hfill
    \begin{subfigure}[b]{0.48\textwidth}
        \centering
        \includegraphics[width=\textwidth]{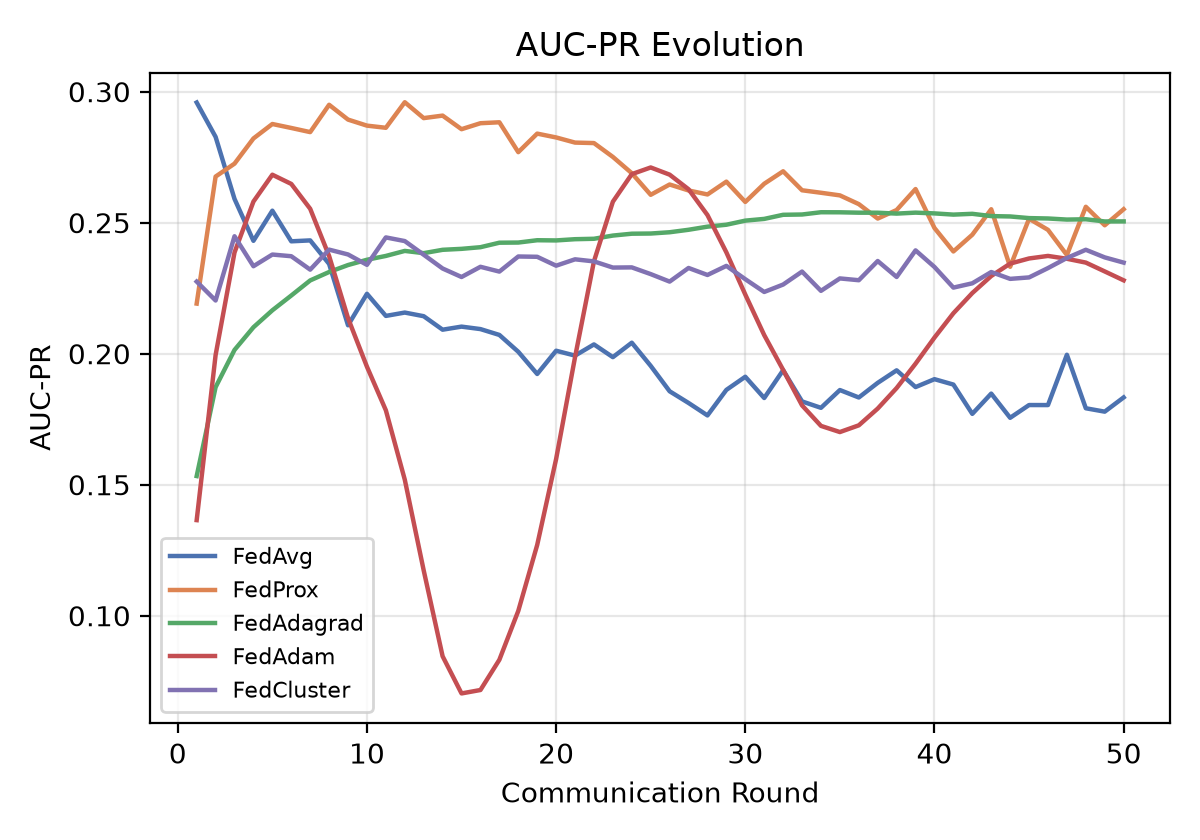}
        \caption{AUC-PR Evolution}
        \label{fig:evo_auc_pr}
    \end{subfigure}
    \vskip\baselineskip
    \begin{subfigure}[b]{0.48\textwidth}
        \centering
        \includegraphics[width=\textwidth]{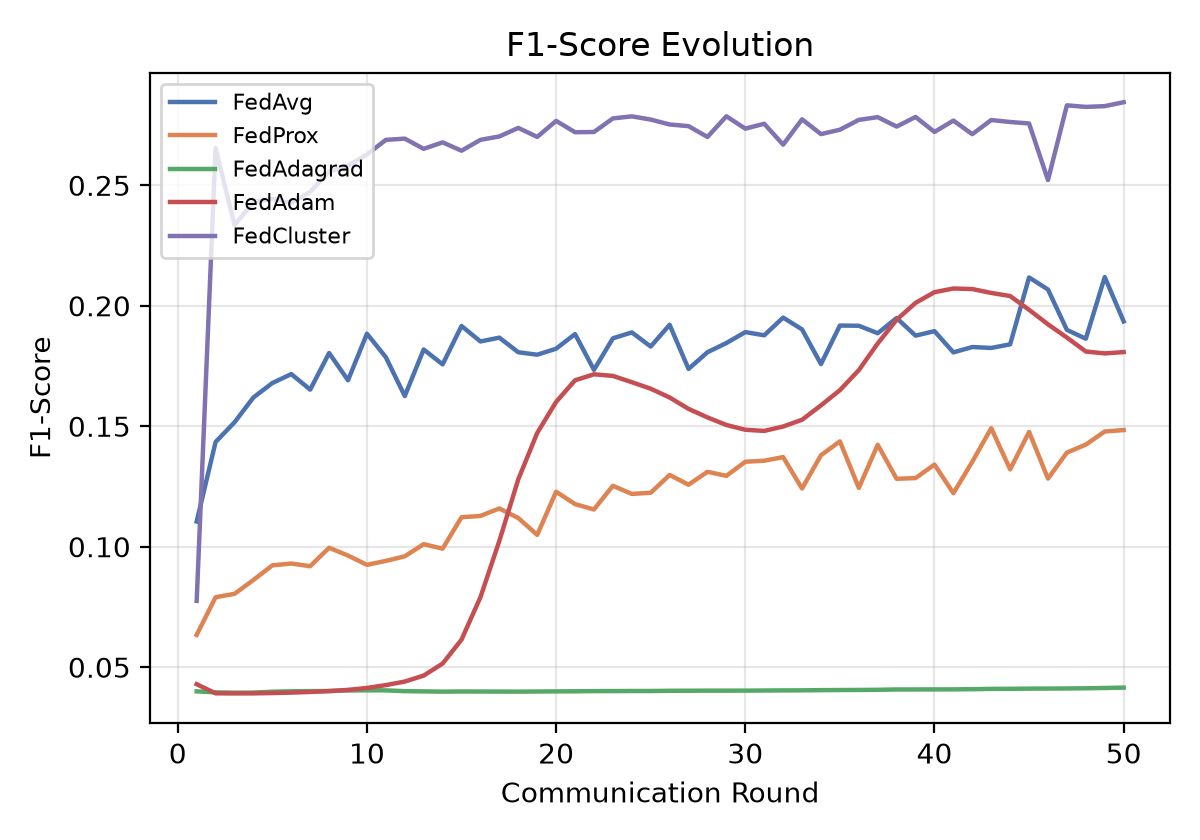}
        \caption{F1-Score Evolution}
        \label{fig:evo_f1_score}
    \end{subfigure}
    \hfill
    \begin{subfigure}[b]{0.48\textwidth}
        \centering
        \includegraphics[width=\textwidth]{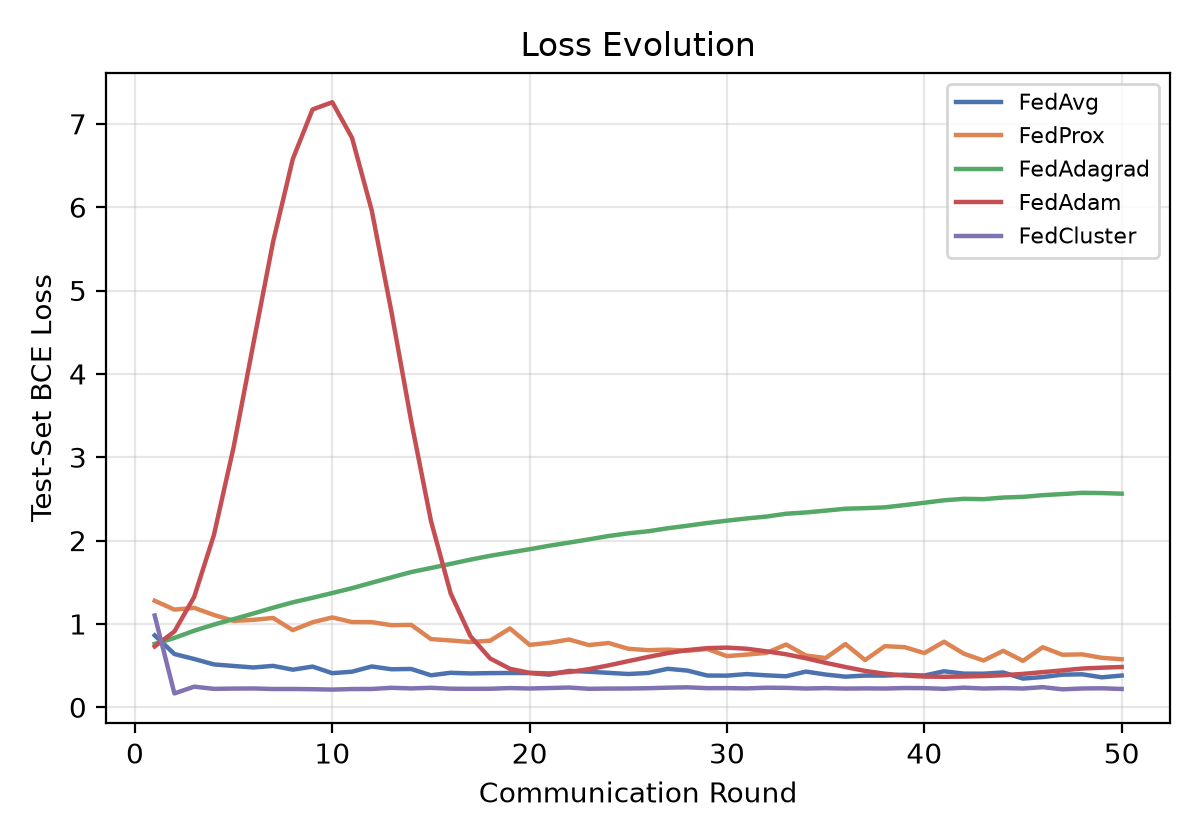}
        \caption{Test-Set Loss Evolution}
        \label{fig:evo_loss}
    \end{subfigure}
    \caption{Evolution of the primary evaluation metrics over 50 communication rounds for the five evaluated federated learning strategies. (a) AUC-ROC, (b) AUC-PR, (c) F1-Score (at a fixed 0.5 threshold), and (d) test-set binary cross-entropy loss.}
    \label{fig:performance_evolution}
\end{figure*}

\begin{figure*}[ht]
    \centering
    \begin{subfigure}[b]{0.32\textwidth}
        \centering
        \includegraphics[width=\textwidth]{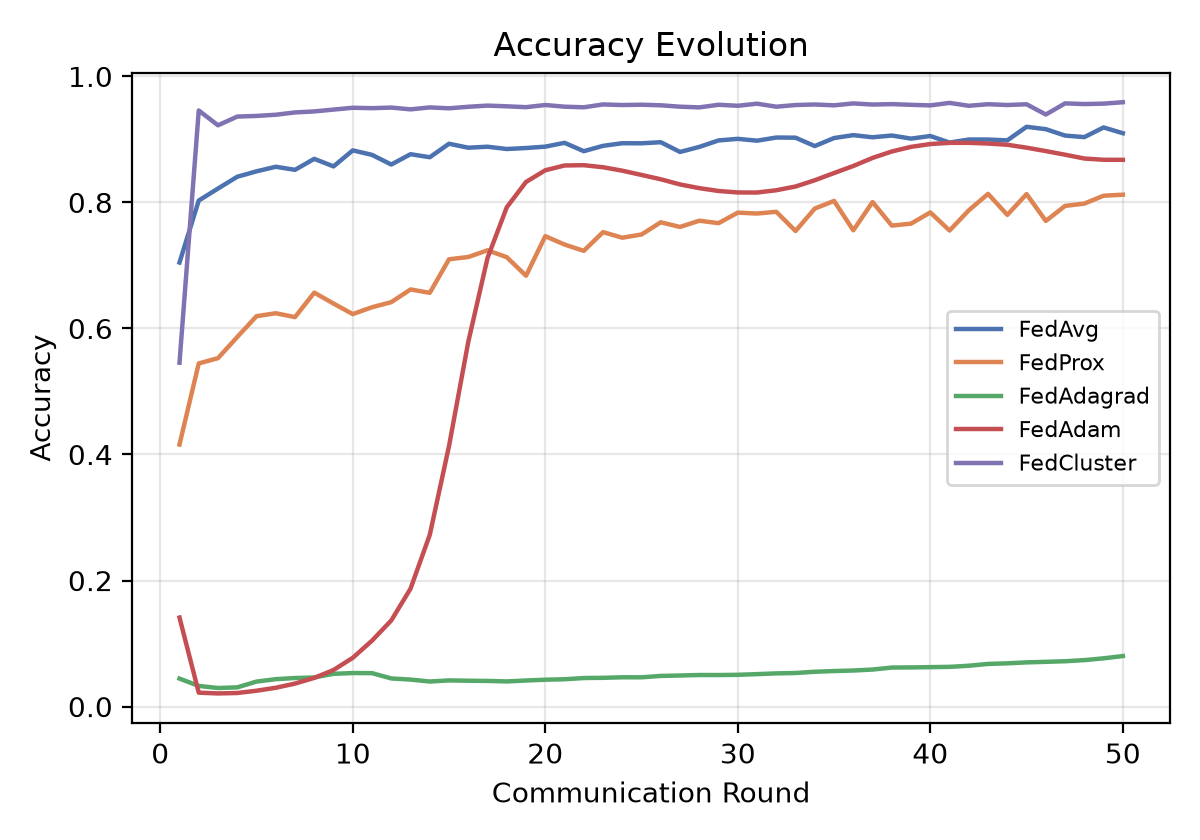}
        \caption{Accuracy Evolution}
        \label{fig:evo_accuracy}
    \end{subfigure}
    \hfill
    \begin{subfigure}[b]{0.32\textwidth}
        \centering
        \includegraphics[width=\textwidth]{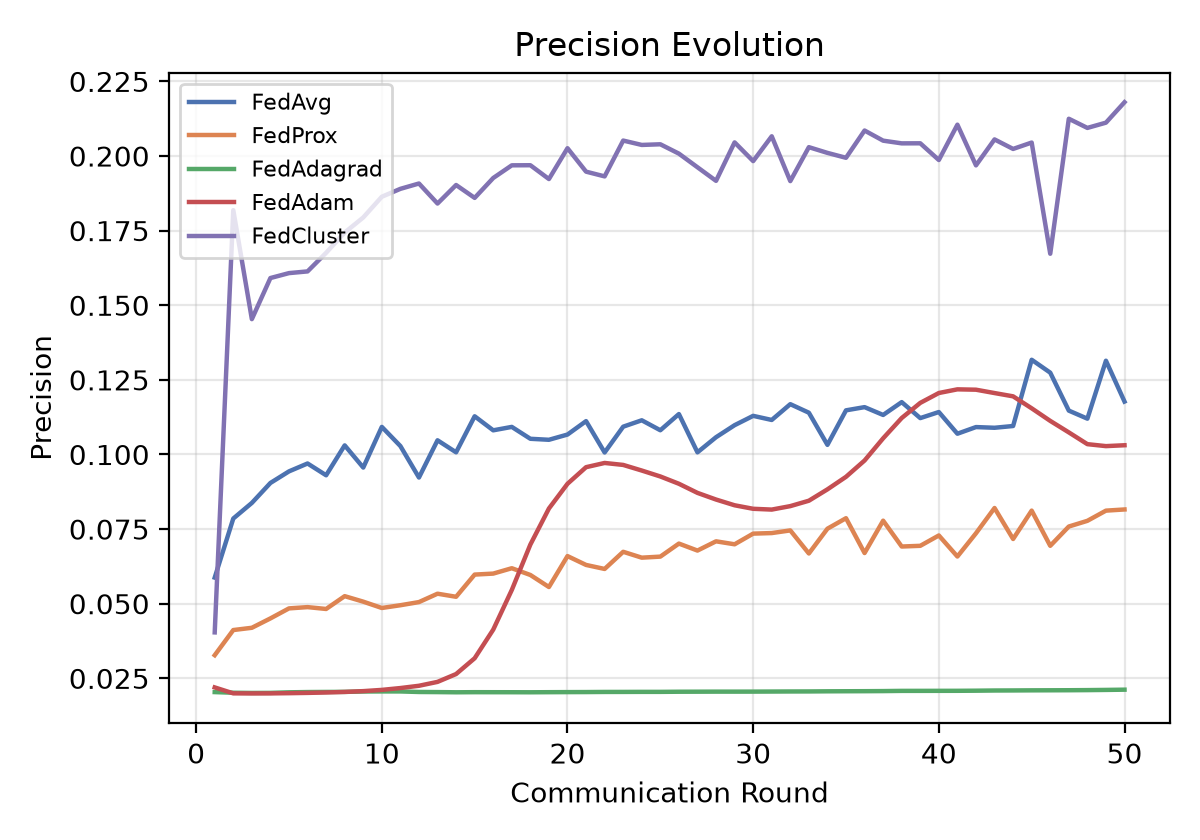}
        \caption{Precision Evolution}
        \label{fig:evo_precision}
    \end{subfigure}
    \hfill
    \begin{subfigure}[b]{0.32\textwidth}
        \centering
        \includegraphics[width=\textwidth]{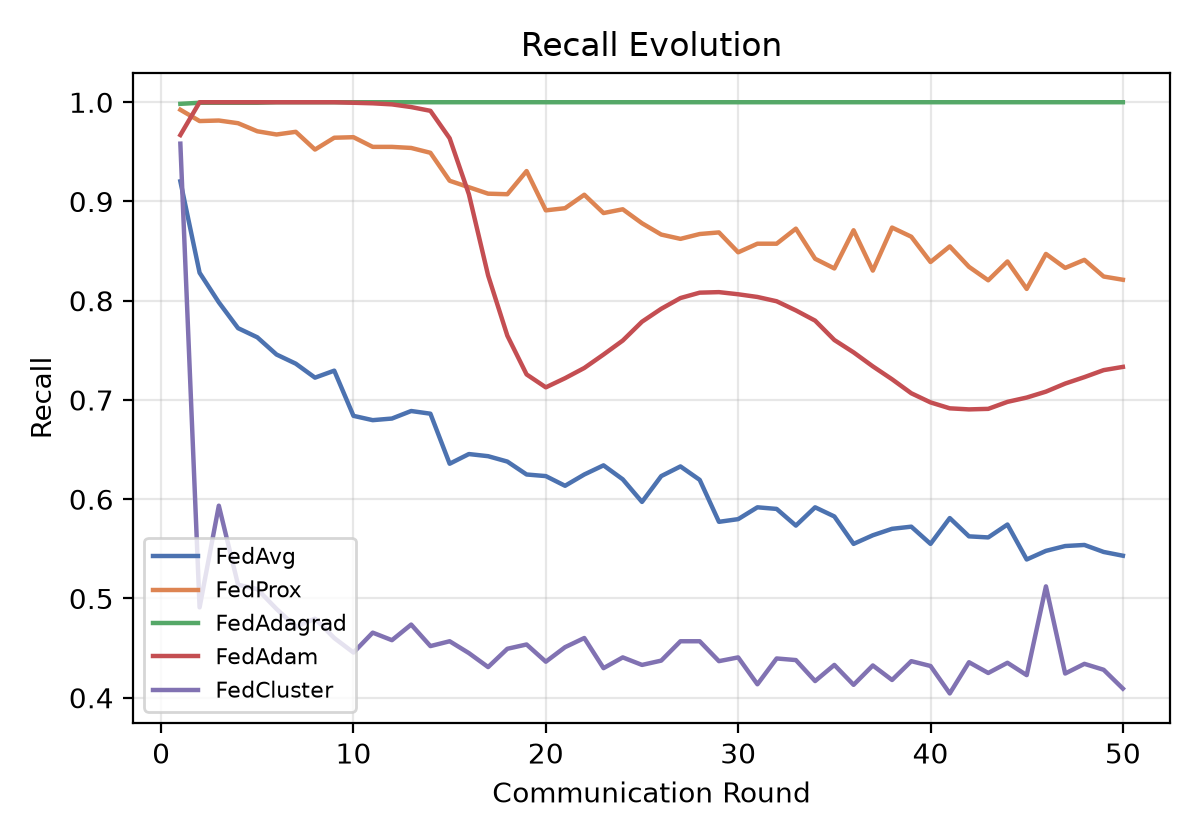}
        \caption{Recall Evolution}
        \label{fig:evo_recall}
    \end{subfigure}
    \caption{Evolution of the secondary, fixed-0.5-threshold metrics over 50 communication rounds. (a) Accuracy, (b) Precision, and (c) Recall.}
    \label{fig:secondary_evolution}
\end{figure*}

The learning behavior of each FL strategy was tracked over 50 communication rounds. Figure~\ref{fig:performance_evolution} illustrates the evolution of our primary, threshold-independent metrics -- AUC-ROC and AUC-PR -- alongside F1-Score (at a fixed 0.5 threshold, per Section~\ref{sec:eval_metrics}) and test-set loss, the four quantities most informative about convergence. Figure~\ref{fig:secondary_evolution} shows the secondary, threshold-dependent metrics (Accuracy, Precision, Recall) at the same fixed 0.5 threshold. The calibrated-threshold numbers reported in Table~\ref{tab:final_results} are computed only once, after round 50.

The AUC-ROC and F1 trajectories tell two related but distinct stories, and the gap between them is itself informative. FedProx has the highest AUC-ROC throughout: it starts at 0.907, peaks at 0.917 around round 4, and settles at 0.898 by round 50 -- the best final value of any strategy despite a gentle decline in the second half of training. FedAdagrad shows the opposite shape: it starts lowest of all five (0.818) and climbs to 0.898 by round 21, then holds essentially flat for the remaining rounds (round-to-round AUC-ROC standard deviation of 0.0021 over rounds 30-50, the lowest of the five) -- a clear, large ranking-quality improvement that its F1 curve barely reflects (F1 moves only from 0.040 to 0.042 over all 50 rounds, and actually peaks at round 1), since F1 at a fixed 0.5 threshold cannot see a model whose probability outputs are shifting but staying on the same side of that threshold. FedAvg and FedCluster both start with the highest AUC-ROC of any strategy at round 1 (0.909 and 0.896, respectively) and \emph{decline} over training (to 0.856 and 0.865), even as their F1 scores rise -- their thresholded decisions are improving while their underlying ranking quality erodes, the same dissociation as FedAdagrad but in the opposite direction. FedAdam is the most visibly unstable: AUC-ROC rises quickly to an early peak (0.909 at round 5), collapses to 0.816 by round 15 -- coinciding almost exactly with its loss spike (peaking at 7.26 around round 10, described below) -- and then recovers unevenly to 0.887 by round 50.

The final calibrated-threshold numbers (Table~\ref{tab:final_results}) show a clear precision/recall trade-off that does not track the AUC-ROC/AUC-PR ranking: FedAdagrad has the highest recall (0.538) of any strategy but the lowest precision (0.149), consistent with its comparatively poor AUC-PR; FedCluster instead achieves the highest precision (0.252) at the cost of one of the lowest recalls (0.319), a combination that also gives it the highest F1 (0.280) of the five. FedProx sits between these extremes (precision 0.193, recall 0.470) and lands the second-highest F1 (0.273), narrowly behind FedCluster and just ahead of FedAdam (0.270) -- so the fixed-threshold F1 ranking is led by a different strategy than the AUC-ROC/AUC-PR ranking, which is exactly why we treat the threshold-independent metrics as primary.

FedAdam's loss spikes sharply, peaking at 7.26 around round 10 -- roughly an order of magnitude above every other strategy's loss at that point -- before recovering by round 18; this transient blow-up coincides almost exactly with the round-15 trough in its AUC-ROC curve, confirming that the model's predicted probabilities became severely miscalibrated at the same time its ranking ability collapsed, consistent with server-side momentum temporarily overshooting during a period of high client disagreement. FedAdagrad again shows the opposite pathology to its AUC-ROC trend: a smooth, monotonic loss \emph{increase} from round 1 (0.76) to round 50 (2.56) even while its AUC-ROC climbs from 0.82 to 0.89 over the same period -- the model's raw confidence calibration degrades steadily across training even as its ability to rank patients by risk genuinely improves, a sign that Adagrad's decaying per-parameter learning rate is pushing probability outputs away from being interpretable as calibrated risk while their relative ordering keeps getting better. FedCluster and FedAvg both end with substantially \emph{lower} loss than they started (FedCluster: 1.10 to 0.22; FedAvg: 0.86 to 0.38), the clearest loss-convergence of the five strategies, even though both also show a declining AUC-ROC over the same rounds -- underscoring, across three different strategies, that loss and ranking-quality metrics can move independently, or even in opposite directions, under non-IID FL, and that neither should be monitored alone.

\subsection{Final Performance Metrics and Efficiency}
\label{sec:final_metrics}
The final performance metrics -- AUC-ROC and AUC-PR (threshold-independent, our primary metrics per Section~\ref{sec:eval_metrics}), plus F1-Score, Accuracy, Precision, and Recall at each strategy's own validation-calibrated decision threshold, each as mean $\pm$ standard deviation over five seeds -- are summarized in Table~\ref{tab:final_results}. By AUC-ROC, our primary metric, FedProx is the top-performing FL strategy ($0.897 \pm 0.002$), ahead of FedAdagrad (0.890), FedAdam (0.885), FedCluster (0.873), and FedAvg (0.857); paired $t$-tests confirm that FedProx's AUC-ROC advantage is statistically significant against every other strategy (all $p < 0.05$; Table~\ref{tab:significance}). FedProx also attains the highest mean AUC-PR (0.230), followed by FedCluster (0.221), FedAdagrad (0.202), FedAdam (0.192), and FedAvg (0.177) -- but on this metric its lead is significant only over FedAdam and FedAvg ($p \leq 0.001$), not over FedAdagrad ($p = 0.074$) or FedCluster ($p = 0.199$), so FedProx is best in mean AUC-PR yet statistically indistinguishable from the two next-closest strategies. The calibrated-threshold F1-score tells yet another story: it is led narrowly by FedCluster (0.280) over FedProx (0.273), FedAdam (0.270), FedAvg (0.255), and FedAdagrad (0.232), so no single strategy dominates all three metrics. Unlike in earlier, unverified drafts of this experiment, no strategy collapses to a degenerate (zero-recall or all-positive) solution -- every strategy converges to a non-trivial classifier, and where FedProx leads, its advantage is a matter of degree (0.007-0.040 AUC-ROC over the other four strategies) rather than a dramatic gap.

\begin{table*}[h!]
\centering
\caption{Final performance metrics for each federated learning strategy after 50 rounds, reported as mean $\pm$ standard deviation over five independent random seeds. AUC-ROC/AUC-PR are threshold-independent (our primary metrics); the remaining four are at each strategy's validation-calibrated threshold. The best mean AUC-ROC, AUC-PR, and F1-score are in bold. Paired significance tests against FedProx are reported in Table~\ref{tab:significance}; execution times in Figure~\ref{fig:execution_time}.}
\label{tab:final_results}
\centering
\begin{tabular}{@{}lcccccc@{}}
\toprule
\textbf{Strategy} & \textbf{AUC-ROC} & \textbf{AUC-PR} & \textbf{F1-Score} & \textbf{Accuracy} & \textbf{Precision} & \textbf{Recall} \\ \midrule
\textbf{FedProx} & \textbf{0.897 $\pm$ 0.002} & \textbf{0.230 $\pm$ 0.019} & 0.273 $\pm$ 0.015 & 0.950 $\pm$ 0.005 & 0.193 $\pm$ 0.018 & 0.470 $\pm$ 0.019 \\
FedAdagrad & 0.890 $\pm$ 0.005 & 0.202 $\pm$ 0.044 & 0.232 $\pm$ 0.019 & 0.929 $\pm$ 0.013 & 0.149 $\pm$ 0.017 & 0.538 $\pm$ 0.049 \\
FedAdam & 0.885 $\pm$ 0.004 & 0.192 $\pm$ 0.022 & 0.270 $\pm$ 0.011 & 0.953 $\pm$ 0.004 & 0.196 $\pm$ 0.013 & 0.437 $\pm$ 0.027 \\
FedCluster & 0.873 $\pm$ 0.008 & 0.221 $\pm$ 0.023 & \textbf{0.280 $\pm$ 0.015} & 0.968 $\pm$ 0.004 & 0.252 $\pm$ 0.023 & 0.319 $\pm$ 0.037 \\
FedAvg & 0.857 $\pm$ 0.006 & 0.177 $\pm$ 0.013 & 0.255 $\pm$ 0.018 & 0.964 $\pm$ 0.003 & 0.216 $\pm$ 0.020 & 0.315 $\pm$ 0.026 \\ \bottomrule
\end{tabular}
\end{table*}

\begin{table}[h!]
\centering
\caption{Paired $t$-tests of FedProx against every other federated strategy on the two primary metrics, over the five seeds (each seed uses an identical held-out test set across strategies, so the samples are paired). $\Delta$ is the mean advantage of FedProx; $p$ is the two-sided paired-$t$ $p$-value.}
\label{tab:significance}
\begin{tabular}{@{}lcccc@{}}
\toprule
 & \multicolumn{2}{c}{\textbf{AUC-ROC}} & \multicolumn{2}{c}{\textbf{AUC-PR}} \\
\cmidrule(lr){2-3}\cmidrule(lr){4-5}
\textbf{FedProx vs.} & $\Delta$ & $p$ & $\Delta$ & $p$ \\ \midrule
FedAdagrad & +0.007 & 0.019 & +0.028 & 0.074 \\
FedAdam    & +0.012 & 0.003 & +0.038 & 0.001 \\
FedCluster & +0.024 & 0.001 & +0.009 & 0.199 \\
FedAvg     & +0.040 & $<$0.001 & +0.053 & $<$0.001 \\ \bottomrule
\end{tabular}
\end{table}

\subsubsection{Per-Client Performance Breakdown}
Following the precedent of \citet{harutyunyan2019multitask}, whose Table 2 reports AUC-ROC broken down by each phenotype's individual prevalence, Table~\ref{tab:per_client_auc} reports the best FL strategy's (FedProx) global model performance evaluated separately on each client's own local, natural-distribution test partition, alongside that client's test-set mortality prevalence. Because the five care units span mortality rates from near-zero (Labor \& Delivery) to almost 3\% (Medicine, Table~\ref{tab:data_partitioning}), this decomposition tests whether the federated global model serves all clients comparably or is implicitly dominated by its largest contributor, the Emergency Department.

\begin{table*}[h!]
\centering
\caption{FedProx global model performance (single representative seed) evaluated separately on each client's own local, natural-distribution test partition (cf. Harutyunyan et al.'s Table 2, which breaks down AUC-ROC by per-phenotype prevalence). Labor \& Delivery has only a single positive example in its test partition, so its ranking metrics ($^{\dagger}$) are not meaningful and are omitted.}
\label{tab:per_client_auc}
\begin{tabular}{@{}lccccc@{}}
\toprule
\textbf{Client (Care Unit)} & \textbf{Test $N$} & \textbf{Prevalence (\%)} & \textbf{AUC-ROC} & \textbf{AUC-PR} & \textbf{F1} \\ \midrule
Emergency Department & 74,216 & 2.29 & \textbf{0.902} & \textbf{0.275} & \textbf{0.313} \\
Medicine/Cardiology & 2,326 & 1.55 & 0.900 & 0.161 & 0.218 \\
Discharge Lounge & 9,834 & 0.32 & 0.876 & 0.152 & 0.073 \\
Medicine & 2,280 & 3.29 & 0.809 & 0.193 & 0.292 \\
Labor \& Delivery & 3,952 & 0.03 & --$^{\dagger}$ & --$^{\dagger}$ & --$^{\dagger}$ \\ \bottomrule
\end{tabular}
\end{table*}

The federated global model does not serve all five clients equally well, which is itself a direct empirical consequence of the non-IID partition described in Section~\ref{sec:non_iid_partitioning}. The two best-served clients, Emergency Department and Medicine/Cardiology, sit at essentially the same AUC-ROC (0.902 and 0.900) despite very different local prevalences (2.29\% vs 1.55\%) and a 32-fold difference in test-set size (74,216 vs 2,326 admissions), while Discharge Lounge (0.876) and, most conspicuously, Medicine (0.809) fall well below them. Medicine has both the highest local mortality prevalence (3.29\%) of any evaluable client and the lowest AUC-ROC -- the opposite of what one would expect if performance simply tracked prevalence -- while also being one of the two smallest clients by volume, suggesting the global model, aggregated with Emergency Department contributing the large majority of training rows (Table~\ref{tab:hyperparameters}), underfits this particular small, clinically distinct client population specifically, rather than small clients in general (Medicine/Cardiology, comparably small, is not similarly disadvantaged). Consistent with AUC-PR's known prevalence-sensitivity~\cite{davis2006relationship}, Discharge Lounge's AUC-PR (0.152) is the lowest of the evaluable clients despite a mid-range AUC-ROC, since ranking quality alone cannot overcome a positive rate of only 0.32\% when precision is computed at any fixed operating point. Labor \& Delivery, with only a single positive example in its test partition, cannot be meaningfully evaluated. This decomposition is only possible because each client retains its own natural-distribution test partition (Section~\ref{sec:non_iid_partitioning}); the pooled global test-set numbers in Table~\ref{tab:final_results} would average over exactly this heterogeneity and hide it.

\begin{figure}[h!]
    \centering
    \includegraphics[width=\columnwidth]{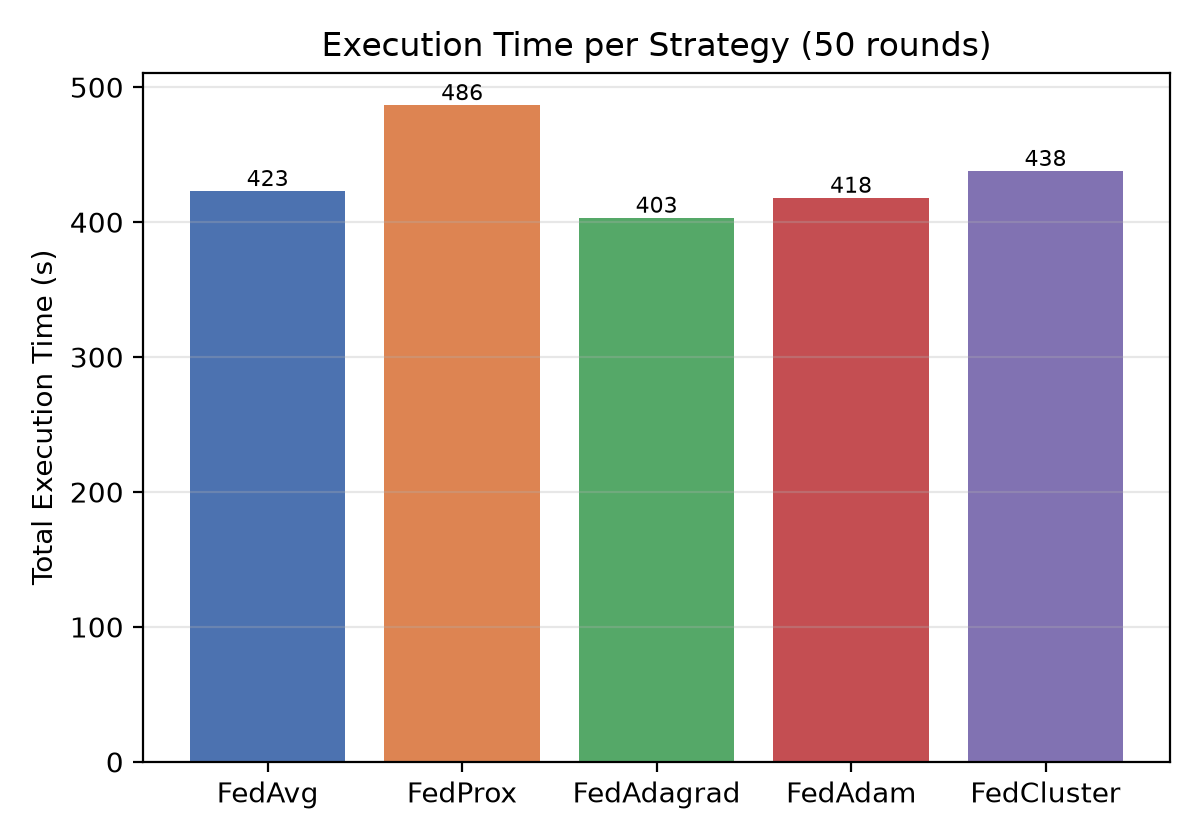}
    \caption{Comparison of total execution time for each FL strategy over 50 rounds.}
    \label{fig:execution_time}
\end{figure}

Regarding computational efficiency, Figure~\ref{fig:execution_time} illustrates the total runtime required for each experiment. FedAdagrad completed the 50 rounds fastest (403.0\,s), and FedProx remained the most computationally intensive (486.4\,s), consistent with the added cost of computing the proximal term at every local update; the remaining three strategies fall within a comparatively narrow 418-438\,s band. Because these runs execute sequentially on shared consumer hardware rather than in a controlled, isolated benchmark environment, these timings should be read as indicative of relative cost rather than a precise benchmark.

\subsection{The Cost of Privacy: Analyzing the Impact of Differential Privacy}
\label{sec:dp_cost}

To further explore the practical implications of deploying federated models in healthcare, an additional experiment was conducted to quantify the impact of incorporating a formal privacy guarantee, Differential Privacy (DP), into our best-performing strategy, FedProx. We implement per-example DP-SGD via Opacus~\cite{opacus2021}, clipping and noising the gradient of each training example before aggregation, rather than a client-level mechanism that would clip and noise each client's entire model update only once per round. This choice matters for privacy accounting: with only 5 clients participating in each round, a client-level mechanism gets essentially no subsampling amplification and yields an unusably large $\varepsilon$ regardless of the noise scale, whereas per-example DP-SGD is amplified by each client's hundreds-to-hundreds-of-thousands of individual training rows. We use noise multiplier $\sigma=1.5$ and gradient-clipping norm $C=1.0$, both standard Opacus defaults, and compute each client's $(\varepsilon, \delta{=}10^{-5})$ guarantee via RDP accounting over the full 50-round, 3-local-epoch schedule. The objective was to measure the trade-offs between privacy, model utility, and computational cost. The results are presented in Table~\ref{tab:dp_comparison}.

\begin{table}[h!]
\centering
\caption{Performance and Efficiency Comparison of FedProx with and without Differential Privacy (DP).}
\label{tab:dp_comparison}
\resizebox{\columnwidth}{!}{%
\begin{tabular}{@{}lcccc@{}}
\toprule
\textbf{Strategy} & \textbf{AUC-ROC} & \textbf{AUC-PR} & \textbf{F1} & \textbf{Time (s)} \\ \midrule
No DP & 0.8976 & 0.2551 & 0.2947 & 486.4 \\
With DP ($\sigma=1.5$) & 0.8963 & 0.2151 & 0.2304 & 2012.3 \\ \bottomrule
\end{tabular}%
}
\end{table}

Per-client privacy budgets at this operating point range from $\varepsilon=1.19$ (Emergency Department, by far the largest client and therefore the strongest subsampling amplification) to $\varepsilon=9.12$ (Labor \& Delivery, which ends up with the smallest training partition after resampling, since it is the one client SMOTE-Tomek cannot enlarge); the remaining clients fall at $\varepsilon=3.64$ (Discharge Lounge), $\varepsilon=8.75$ (Medicine/Cardiology), and $\varepsilon=8.85$ (Medicine). All five fall within a range widely considered meaningful in the applied DP literature, in contrast to the $10^5$-$10^6$ range a naive client-level mechanism would produce under the same 5-client, no-subsampling setting.

The results show a smaller, more specific utility cost than a purely client-level DP mechanism would yield, and one that is easy to miss if only a single metric is examined. AUC-ROC is nearly unaffected by DP-SGD (0.898 to 0.896, a difference of 0.0013) -- the model's underlying ability to rank patients by mortality risk is essentially preserved. AUC-PR (0.255 to 0.215, a 15.7\% relative reduction) and F1 (0.295 to 0.230, a 21.8\% relative reduction) fall more, and the calibrated operating point shifts substantially: recall increases from 0.443 to 0.674 while precision falls from 0.221 to 0.139, and accuracy drops from 0.958 to 0.910. This pattern -- ranking quality preserved, operating-point precision/recall trade-off shifted -- is consistent with per-example gradient clipping and Gaussian noise injection perturbing the model's absolute confidence calibration more than its relative ordering of patients: DP-SGD's two core operations (clipping each example's gradient contribution, then adding calibrated noise to the clipped, summed gradient before the optimizer step) distort the fine-grained probability estimates a fixed threshold depends on, without as strongly disturbing which patients are ranked as higher- versus lower-risk relative to one another, in exchange for the formal $(\varepsilon,\delta)$ guarantee reported above.

Implementing DP also introduced substantial computational overhead: total training time increased from 486.4 seconds to 2012.3 seconds, a 4.1$\times$ increase, driven by Opacus's need to compute and clip a gradient for every individual training example rather than a single averaged mini-batch gradient -- a substantially larger overhead than the aggregate proximal-term cost that separates FedProx from FedAvg in Table~\ref{tab:final_results}. Our findings suggest that while per-example DP-SGD offers a materially better privacy/utility trade-off than the client-level alternative described above (single-digit-to-low-double-digit $\varepsilon$ instead of $10^5$-$10^6$) and preserves ranking quality (AUC-ROC) almost entirely, it remains neither free nor negligible at any fixed operating point: institutions must budget for a real shift in the precision/recall trade-off, a real drop in AUC-PR and F1, and a multi-fold increase in training time when deploying formally private federated learning for real-world clinical applications.

\section{Discussion}
\label{sec:discussion}

The primary objective of the study was to conduct a rigorous comparison of five distinct federated learning strategies in a realistic, non-IID, and imbalanced clinical setting for in-hospital mortality prediction. Our results indicate that the choice of aggregation strategy has a real, and by our primary metrics unambiguous, impact on model performance under this cohort's specific partitioning: FedProx leads every other strategy on our primary metric AUC-ROC (0.897, next-best 0.890; the lead is statistically significant against all four by paired $t$-test) and attains the highest mean AUC-PR (0.230, next-best 0.221), while on the calibrated-threshold F1-score it is narrowly edged out by FedCluster (0.280 vs 0.273); none of the five collapses to a degenerate solution. We interpret FedProx's clear and statistically significant AUC-ROC lead as meaningful evidence for the value of algorithms that explicitly manage statistical heterogeneity; the round-by-round trajectories (Section~\ref{sec:eval_metrics}, Figure~\ref{fig:performance_evolution}) show that this ranking need not coincide with the fixed-threshold one -- several strategies' AUC-ROC and F1 curves move in opposite directions across training, and at the final round the best strategy by F1 (FedCluster) is not the one that is best by AUC-ROC/AUC-PR (FedProx), a distinction a fixed-threshold metric alone would obscure.

FedProx's edge is plausibly attributable to its mechanism for mitigating "client drift": in non-IID environments, local models trained on client-specific data distributions tend to diverge from one another, and the proximal term in FedProx's local objective penalizes large deviations from the global model, constraining local updates toward a more stable, shared trajectory. Its AUC-ROC trajectory (Figure~\ref{fig:evo_auc_roc}) reflects this directly -- it is at or near the top of all five strategies from round 1 onward and never drops below 0.89 after its early peak, the most consistently strong ranking performance of the five, and a lead that the paired $t$-tests confirm is statistically significant over every other strategy (Table~\ref{tab:significance}). This stability also shows up in the calibrated-threshold numbers: rather than sitting at either extreme of the precision/recall trade-off the other four strategies are pulled toward (Section~\ref{sec:final_metrics}), FedProx lands closer to the middle (precision 0.193, recall 0.470), giving it the second-highest F1 (0.273) -- narrowly behind FedCluster (0.280) -- even though its advantage on the threshold-independent metrics does not translate into a lead on the fixed-threshold F1.

The two server-side adaptive strategies, FedAdagrad and FedAdam, tell a more nuanced story than a simple pass/fail, one that is only fully visible by tracking AUC-ROC, F1, and loss together rather than any single metric. FedAdam's AUC-ROC trajectory (Figure~\ref{fig:evo_auc_roc}) rises quickly to an early peak (0.909 at round 5), collapses to 0.816 by round 15 in lockstep with its loss spike (Figure~\ref{fig:evo_loss}, peaking at 7.26 around round 10), and recovers unevenly thereafter -- consistent with the server-side momentum term amplifying, rather than damping, oscillations introduced by highly divergent client updates during a period of acute client disagreement. Two properties of the update rule (Section~\ref{sec:fl_strategies}) make this amplification more likely here than in the cross-device settings FedAdam was originally designed for: with only five, highly non-IID clients per round, the pseudo-gradient $\Delta_t$ it aggregates is far noisier round-to-round than an average over hundreds or thousands of clients would be, and, following the update rule of \cite{reddi2021adaptivefederatedoptimization} exactly, $m_t$ and $v_t$ receive no bias correction and $v_t$ is initialized at zero with a slow decay ($\beta_2=0.99$), so during the first several rounds the denominator $\sqrt{v_t}+\tau$ systematically underestimates the true update magnitude and the effective step size is inflated precisely when the aggregated direction is least trustworthy -- consistent with the loss rising monotonically from round 1 through the round-10 peak before $v_t$ catches up and the effective step shrinks back down. FedAdagrad shows the opposite failure mode: its AUC-ROC climbs steadily and substantially (0.82 to 0.89) while its F1 barely moves at all (0.040 to 0.042, in fact peaking at round 1) and its loss \emph{rises} monotonically throughout training (0.76 to 2.56) -- the model's ability to rank patients by risk keeps improving even as its probability calibration steadily degrades, a dissociation that a single metric could not have revealed and that we attribute to Adagrad's decaying per-parameter learning rate pushing raw probability outputs away from being well-calibrated even as their relative ordering improves. At its calibrated threshold, this translates into the highest recall of any strategy (0.538) but also the lowest precision (0.149) and the lowest F1 (0.232) -- FedAdagrad's real ranking improvement is being spent on catching more true positives, at a steep cost in false positives.

The baseline FedAvg is, perhaps surprisingly, weakest of the five on both primary metrics (AUC-ROC 0.857, AUC-PR 0.177) despite starting round 1 with the single highest AUC-ROC of any strategy (0.909) -- its ranking quality \emph{erodes} over the 50 rounds even as its F1 rises, the same loss-vs-ranking dissociation seen in FedAdagrad but running in the opposite direction; at its calibrated threshold it also has the lowest recall of the five (0.315), alongside a mid-range precision (0.216). FedCluster, in contrast, presents the most favorable calibrated-threshold profile of the five: the highest precision (0.252), the best F1 (0.280), the second-highest AUC-PR (0.221, behind only FedProx), and by far the lowest, most stable loss throughout training (falling from 1.10 to 0.22). Its one clear weakness is AUC-ROC (0.873, fourth of the five): its \emph{ranking} quality lags FedProx, FedAdagrad, and FedAdam even though its thresholded decisions and precision/recall balance are the best-calibrated of the group. This combination -- excellent precision, F1, and loss convergence, but middling AUC-ROC -- is consistent with the DBSCAN outlier-filtering step sometimes discarding a client whose update was directionally noisy but still informative for the minority (mortality) class, trading away some ranking consistency for a more conservative, better-calibrated consensus model.

Our findings have practical implications for deploying FL in healthcare, though more measured ones than a simple "FL wins" story: for networks of institutions with diverse patient populations, FedProx is a reasonable default among the federated strategies evaluated here, offering the best AUC-ROC (significantly so) and the best mean AUC-PR, together with the best balance of stability and predictive performance, without a specific weakness like FedAdagrad's early plateau, FedAdam's transient instability, or FedCluster's weaker ranking quality. But the comparison against centralized baselines later in this section (Table~\ref{tab:final_results_centralized}) also shows that, on this cohort, centralized training modestly outperforms every federated strategy on both primary metrics -- so the choice to federate should be justified primarily by the privacy and data-governance requirements that motivate FL in the first place, not by an expectation of a predictive-performance advantage.

A crucial aspect of our analysis was to quantify the performance of the federated models in comparison to traditional centralized baselines, evaluated on the identical held-out global test set used for the federated strategies (same rows, same natural class distribution) to ensure the comparison is like-for-like. We evaluated three distinct centralized strategies: a naive approach without imbalance handling, a model trained on globally resampled data using SMOTE-Tomek, and a model with a class-weighted loss function, each also calibrated to its own F1-maximizing decision threshold on the validation set. As shown in Table~\ref{tab:final_results_centralized}, the best centralized approach (naive, no explicit imbalance handling at all) reaches AUC-ROC 0.929 and AUC-PR 0.312, ahead of FedProx's 0.897 and 0.230 by 0.032 and 0.082 respectively (the latter a 36\% relative gap) -- a real, if moderate, centralized advantage on both primary metrics, and one that the paired $t$-tests confirm is statistically significant ($p<0.001$ for both AUC-ROC and AUC-PR). Somewhat counter-intuitively, the two centralized strategies that explicitly address class imbalance (global SMOTE and weighted loss) both do \emph{worse} than the naive baseline on AUC-ROC (0.886 and 0.927, respectively) despite the naive model never seeing a rebalanced training distribution -- consistent with imbalance-handling techniques optimizing for a different point on the precision-recall trade-off (both achieve higher recall than naive) rather than for ranking quality per se. This is the more literature-consistent outcome for centralized-vs-federated comparisons: centralized training, which sees the full, non-partitioned dataset without any of the aggregation or client-drift losses inherent to FL, would be expected to match or modestly exceed federated performance under a fair comparison, and that is what we observe here on both AUC-ROC and AUC-PR. We view this modest, consistent gap -- not the dramatic reversal an earlier, unverified draft of this experiment reported -- as the more credible picture of what decentralization costs on this specific cohort, once the evaluation protocol keeps the class distribution honest and the comparison groups see the same held-out data. Both figures also sit comfortably within the range reported by prior centralized MIMIC-IV mortality studies -- albeit on different cohorts and prevalences: a boosted nonparametric-hazards model for real-time ICU mortality reports AUC-ROC 0.83~\cite{nowroozilarki2021realtime}, and an XGBoost model for in-hospital mortality among acute-ischemic-stroke ICU patients reaches AUC-ROC 0.86~\cite{cummins2025stroke}. Our centralized 0.929 sits at the upper end of, and our federated 0.857--0.897 squarely within, this published performance band, indicating that the privacy-preserving federated setup pays only a modest accuracy cost relative to established centralized benchmarks rather than operating in a degraded regime.

\begin{table*}[h!]
\centering
\caption{Final performance metrics (mean $\pm$ standard deviation over five seeds) for all evaluated strategies, all sharing the identical held-out test set. The best federated model (FedProx) and the overall-best model (Centralized Naive) are highlighted by AUC-ROC.}
\label{tab:final_results_centralized}
\setlength{\tabcolsep}{4pt}
\small
\begin{tabular}{@{}lcccccc@{}}
\toprule
\textbf{Strategy} & \textbf{AUC-ROC} & \textbf{AUC-PR} & \textbf{F1-Score} & \textbf{Accuracy} & \textbf{Precision} & \textbf{Recall} \\ \midrule
\multicolumn{7}{c}{\textit{Federated Learning Strategies}} \\
FedAvg & 0.857 $\pm$ 0.006 & 0.177 $\pm$ 0.013 & 0.255 $\pm$ 0.018 & 0.964 $\pm$ 0.003 & 0.216 $\pm$ 0.020 & 0.315 $\pm$ 0.026 \\
\textbf{FedProx} & \textbf{0.897 $\pm$ 0.002} & \textbf{0.230 $\pm$ 0.019} & 0.273 $\pm$ 0.015 & 0.950 $\pm$ 0.005 & 0.193 $\pm$ 0.018 & 0.470 $\pm$ 0.019 \\
FedAdagrad & 0.890 $\pm$ 0.005 & 0.202 $\pm$ 0.044 & 0.232 $\pm$ 0.019 & 0.929 $\pm$ 0.013 & 0.149 $\pm$ 0.017 & 0.538 $\pm$ 0.049 \\
FedAdam & 0.885 $\pm$ 0.004 & 0.192 $\pm$ 0.022 & 0.270 $\pm$ 0.011 & 0.953 $\pm$ 0.004 & 0.196 $\pm$ 0.013 & 0.437 $\pm$ 0.027 \\
FedCluster & 0.873 $\pm$ 0.008 & 0.221 $\pm$ 0.023 & 0.280 $\pm$ 0.015 & 0.968 $\pm$ 0.004 & 0.252 $\pm$ 0.023 & 0.319 $\pm$ 0.037 \\ \midrule
\multicolumn{7}{c}{\textit{Centralized Baseline Strategies}} \\
\textbf{Centralized (Naive)} & \textbf{0.929 $\pm$ 0.003} & \textbf{0.312 $\pm$ 0.027} & 0.348 $\pm$ 0.014 & 0.971 $\pm$ 0.005 & 0.318 $\pm$ 0.045 & 0.394 $\pm$ 0.048 \\
Centralized (Global SMOTE) & 0.886 $\pm$ 0.012 & 0.235 $\pm$ 0.020 & 0.276 $\pm$ 0.016 & 0.967 $\pm$ 0.002 & 0.246 $\pm$ 0.015 & 0.315 $\pm$ 0.026 \\
Centralized (Weighted Loss) & 0.927 $\pm$ 0.002 & 0.307 $\pm$ 0.023 & 0.351 $\pm$ 0.013 & 0.968 $\pm$ 0.003 & 0.296 $\pm$ 0.029 & 0.436 $\pm$ 0.032 \\ \bottomrule
\end{tabular}
\end{table*}

\subsection{Privacy-Enhancing Technologies and The Cost of Security}

It is essential to note that while FL aims to preserve data privacy by sharing model parameters rather than raw data, the framework itself does not eliminate all security concerns. Seminal studies have shown that it is possible to infer properties of the training data by examining model parameter updates \cite{aono2017privacy}. To counter this, several Privacy-Enhancing Technologies (PETs) can be integrated with FL, most notably Differential Privacy (DP)~\cite{10.1561/0400000042}, Homomorphic Encryption (HE)~\cite{10.1145/2857705.2857731}, and Secure Multi-Party Computation (SMPC) \cite{10.1145/3658644.3690257}.

These methods, however, introduce significant trade-offs between privacy, model utility, and computational efficiency. Our supplementary experiment, detailed in Section~\ref{sec:dp_cost}, provides an empirical measurement of this trade-off for DP. By applying per-example DP-SGD to our best-performing model, FedProx, at a standard operating point ($\sigma=1.5$, $C=1.0$), we observed AUC-ROC essentially unchanged (0.898 to 0.896) but F1 fell from 0.295 to 0.230 (a 21.8\% relative reduction) and AUC-PR fell from 0.255 to 0.215, accompanied by a 4.1$\times$ increase in computational time, in exchange for a per-client privacy budget of $\varepsilon=1.19$-$9.12$ ($\delta=10^{-5}$). This demonstrates that while DP provides a formal, and in this case reasonably tight, privacy guarantee that leaves the model's ranking ability nearly intact, the cost to its calibrated operating point and to training time is real and non-negligible. Other technologies, such as HE, which ensures that only encrypted model parameters are exchanged, and SMPC, which preserves knowledge of client inputs through cryptographic protocols, offer stronger security without the same DP-utility trade-off. However, they are known to impose an even greater penalty on computational and communication resources, further increasing the already considerable training times observed in our study. Further research is needed to understand the complex trade-offs that arise when combining advanced PETs with different FL algorithms in real-world EHR data.

\subsection{Ethical Considerations and Limitations}
While our findings are promising, the real-world deployment of predictive models in clinical settings requires careful ethical consideration. The MIMIC-IV dataset, although extensive, represents a specific patient population, and the performance of our models may not generalize equally across different demographic groups. Future work should include a rigorous fairness analysis to audit for potential algorithmic biases. Furthermore, we emphasize that mortality prediction models should be implemented as decision-support tools to augment, not replace, human clinical judgment. Federated Learning promotes responsible data governance by preserving patient privacy, a critical ethical prerequisite for multi-institutional collaboration in medical research.

\section{Conclusion and Future Work}
\label{sec:conclusion_future_work}

\subsection{Conclusion}
The study conducted a comprehensive comparative analysis of five distinct federated learning strategies for the critical task of in-hospital mortality prediction on a large, real-world intensive care dataset characterized by statistical heterogeneity and class imbalance. Our objective was to evaluate the performance, stability, and efficiency of baseline, regularized, adaptive, and clustering-based FL algorithms in a simulated yet realistic non-IID environment.

The empirical results demonstrate that the choice of aggregation strategy has a measurable impact on model performance under this cohort's non-IID partition, most clearly visible in AUC-ROC and AUC-PR, our primary metrics (Section~\ref{sec:eval_metrics}): none of the five strategies collapses to a degenerate classifier, and \textbf{FedProx}, which introduces a client-side regularization term, is the strongest of the five on our primary metric AUC-ROC (0.897, significantly so against all four competitors) and on mean AUC-PR (0.230), though on the calibrated-threshold F1-score it is narrowly surpassed by FedCluster (0.280 vs 0.273); the margin over the closest competitor on any one metric is modest, not dramatic. \textbf{FedAdagrad} is the AUC-ROC runner-up (0.890) despite the flattest F1 trajectory of the five, a dissociation traced in Section~\ref{sec:discussion} to steadily improving ranking quality alongside steadily degrading probability calibration -- and, at its calibrated threshold, the highest recall (0.538) and lowest precision (0.149) of any strategy. \textbf{FedAdam} (AUC-ROC = 0.885) shows the most visible instability, with a sharp transient loss spike and AUC-ROC dip around round 10-15. The baseline \textbf{FedAvg} has the weakest AUC-ROC (0.857) of the five despite starting training with the single highest value of any strategy, and \textbf{FedCluster} (AUC-ROC = 0.873) achieves the highest precision (0.252) and the best F1 (0.280) at its calibrated threshold, plus the second-highest AUC-PR and the lowest, most stable loss, but trails on AUC-ROC. Critically, when the same held-out, natural-class-distribution test set is used to evaluate three centralized baselines, the best of them (naive, no explicit imbalance handling) reaches AUC-ROC = 0.929 and AUC-PR = 0.312, significantly ahead of FedProx on both ($p<0.001$) -- centralization retains a real, if moderate, predictive edge on this task once the evaluation protocol is held constant across both paradigms. Breaking FedProx's performance down by client further shows that this single aggregate AUC-ROC of 0.897 masks a 0.809-0.902 range across the evaluable care units, with the model underserving its Medicine client specifically.

The primary contribution of the work is a direct, empirical benchmark, together with two methodological findings that we consider at least as important as the benchmark itself. First, keeping the evaluation test set at the true class distribution, keeping it identical across federated and centralized comparisons, and reporting AUC-ROC/AUC-PR rather than a fixed-threshold metric as primary (following \citet{harutyunyan2019multitask} and \citet{davis2006relationship}) together materially change the conclusions such a benchmark supports: the round-by-round trajectories in Section~\ref{sec:discussion} show AUC-ROC and F1 repeatedly moving in opposite directions within a single strategy (FedAdagrad's ranking quality improving while its thresholded decisions stand still; FedAvg's and FedCluster's ranking quality eroding while their thresholded decisions improve), a distinction a single fixed-threshold metric could not have revealed -- and, at the final round, the strategy that is best by F1 (FedCluster) is not the one that is best by AUC-ROC/AUC-PR (FedProx). Second, aggregate metrics computed on a pooled global test set can mask performance heterogeneity across the very clients an FL system is meant to serve -- our per-client breakdown (Table~\ref{tab:per_client_auc}) shows FedProx's global model ranging from 0.902 AUC-ROC on its best-served client (Emergency Department) down to 0.809 on Medicine across the four evaluable clients, despite a roughly 10-fold range in local prevalence (0.32\% to 3.29\%) and a 33-fold range in test-set size, a within-strategy spread comparable to the gap between the best and worst \emph{strategy} in Table~\ref{tab:final_results}. Our results suggest that regularization-based methods, such as FedProx, remain a more robust choice \emph{among} FL aggregation strategies for non-IID and imbalanced tasks like mortality prediction, but that the decision to federate at all should rest on privacy and data-governance requirements rather than an expectation of superior predictive performance over centralized training. The study also confirms the practical necessity of a per-client fallback for class imbalance handling techniques like SMOTE-Tomek: one of our five clients (Labor \& Delivery) has only a single mortality event in over 20,000 admissions -- so its training partition holds at most one positive example, below SMOTE's minimum neighbor requirement -- and required a class-weighted-loss fallback rather than resampling. Limitations of our work include the experimental setup in a federated environment using a single-source dataset, the fixed number of five clients, and a batch size (512) larger than typical for this scale of data, chosen for computational tractability on commodity hardware (Section~\ref{sec:methodology}).

\subsection{Future Work}
Building upon the study's findings, several avenues for future research can be explored. A natural next step would be to expand the scale of the experiments by increasing the number of clients and exploring different non-IID data partitioning schemes to further test the algorithms' robustness, and to re-run this benchmark with a smaller client batch size on hardware where the resulting compute cost is tractable, to isolate whether the batch-size increase used here materially affects the relative ranking of strategies. Investigating the performance of more advanced deep learning architectures, such as Transformers or Recurrent Neural Networks (RNNs), which can better model the time-series nature of EHR data, presents another promising direction. Conversely, a linear baseline (logistic regression) under the same five aggregation strategies would isolate how much of the reported performance depends on the DNN's non-linear capacity versus the aggregation strategy itself -- our accompanying codebase already includes a logistic-regression model class for this purpose, though it was not run for the results reported here.

Furthermore, future work could explore hybrid federated strategies that combine the stability of FedProx with the adaptive learning capabilities of client-side optimizers. A deeper investigation into FedAdam's oscillatory convergence and FedAdagrad's early plateau in this context could yield valuable insights into their sensitivity to hyperparameters and their interactions with client-side data resampling techniques. Finally, while our work analyzed the utility cost of DP, a crucial future step would be to integrate cryptographic methods, such as Secure Multiparty Computation (SMPC) and Homomorphic Encryption (HE). A formal analysis of the trade-offs these techniques introduce between computational overhead and security guarantees would be essential for designing production-ready, privacy-preserving clinical systems.

\section*{Declarations}

\begin{contributions}
Rodrigo Ronner Tertulino da Silva is the sole author of the paper.
\end{contributions}

\begin{interests}
The author declares no competing interests.
\end{interests}

\begin{acknowledgements}
The author would like to acknowledge the support of the Software Engineering and Automation Research Laboratory (LaPEA), where the research was developed and conducted. The infrastructure and resources provided were essential to completing the work. 
We especially appreciate the MIMIC-IV official team’s efforts to open-source
the database and codes.
\end{acknowledgements}

\begin{funding}
This research was not funded.
\end{funding}

\begin{materials}
All data in the article can be obtained from the MIMIC-IV database (https://physionet.org/content/mimiciv/3.1/), subject to PhysioNet's credentialed-access process. The complete source code used to produce every result in this paper is publicly available at \url{https://github.com/rodrigoronner/FedMIMIC-Mortality-Benchmark}.
\end{materials}

\section*{Use of AI and Language Tools}
During the preparation of this work, the author used Claude (Anthropic, 2025) and Grammarly to improve text fluency, grammar, spelling, and readability. These tools were used exclusively for language refinement. They were not used to generate research content, fabricate data, create citations, perform analysis, or write substantive portions of the manuscript. All suggestions were carefully reviewed, edited, and validated by the author, who takes full responsibility for the accuracy and integrity of all content.

\bibliographystyle{apalike-sol}
\bibliography{refs}

\end{document}